# Study of Human Push Recovery

*Kang Tan*

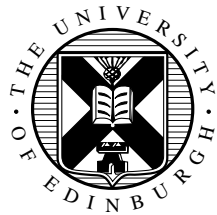

Master of Science
Artificial Intelligence
School of Informatics
University of Edinburgh
2018

# Abstract

Walking and push recovery controllers for humanoid robots have advanced throughout the years while unsolved gaps leading to undesirable behaviours still exist. Because previous studies are mainly pure engineering methods, while the use of data-driven methods has made impressive achievements in the field of control, we set the motivation for exploration of control laws applied by human beings. Successful findings may help fill the gaps in current engineering-based controllers and can potentially improve the performance.

In this thesis, we show our complete design and implementation of a set of experiments to collect and process human data, as well as data analysis for model fitting. Using the processed motion data and force data collected, we export the position, velocity and acceleration of our participants' centres of mass to form a one-dimensional point mass model defined by the direction of pushes. As a result, we find that proportional-derivative (PD) control can describe the underlying control law that people used and different PD gains are set for different phases of a push recovery trial within the scope of our study. Our final PD control fittings have an average root mean square error below 0.1 with above 90% of a trial's data points are taken into account on average. We also explore how different error metrics that the PD control model uses influence its performance but neither of the two metrics we proposed can help improve the performance. Finally, we have statistics on how people switch push recovery strategies based on the their centre of mass properties at the start of the push. We find that the further the centre of mass is from the steady-state point and the higher velocity it has, a person is more likely to make a step for push force compensation.

# Acknowledgements

My sincere thanks to Chris, Daniel and Kai, for their patience and valuable advice throughout this project.

Also, I am much appreciated to my supervisor, Dr. Zhibin Li, who gives a lot of freedom but also crucial supervisions which help me explore the topic without any constrain while not getting lost.

# Declaration

I declare that this thesis was composed by myself, that the work contained herein is my own except where explicitly stated otherwise in the text, and that this work has not been submitted for any other degree or professional qualification except as specified.

(*Kang Tan*)

# Table of Contents

# Chapter 1

# Introduction

This thesis explores the control laws applied by human beings during *push recovery* with a simplified dynamics model and a predefined hypothesis. Before listing the contribution we have made, we will first give the motivation and the goal set for this project. This chapter also includes brief information about the thesis structure. Note that section 1.1 is based on the introduction section of the author's proposal for this project [47].

## 1.1  Motivation

Compared with fixed robotic platforms, mobile robots can be useful in various environments. Wheeled robots have been a kind of such mobile robots being capable of moving on various surface environments. However, specific designs aimed to work well in one arena usually do not generalise well, and wheels become less useful in limited surface area terrain. Therefore, a solution lies in using bipedal or quadrupedal robots being able to walk in multiple environments. However, using inherently wobbly actuated legs leads to difficult control problems since a legged robot's weight is distributed mainly on its upper body which is more less stable in structure, compared with wheels and other supports.

Walking control has become much more advanced throughout the years while real-world implementations are still in the face of inaccuracy and incomplete sensor measurements, environmental dynamics changes and so on. These issues lead to potential failure of a walking plan's execution since usually feed-forward walking controllers are implemented. We call such a problem a disturbance if it occurs during the execution of a walking plan, and disturbances encompass several kinds of potential threats [19]: Pushes, Trips, Slip, Foot Placement Restriction and Collision Avoidance. A push, our interested disturbance, can be defined as a force applied to a point of a subject being able to influence its stability. A push may affect the state of standing, walking, and running. Researches in disturbance rejection have been seeking solutions to all these threats and *push recovery* study, one aspect in this research area to deal with pushes, has become well developed with a number of potential solutions presented.

The majority of push recovery studies in robotics are purely engineering-based methods taking pushes' physical profiles, the robot's kinematics and dynamics into account to develop recovery behaviours. These studies have reached valuable success yet lack still exists such as a humanoid robot reacting to pushes and other abuses in an unnatural way of always keeping its knees bent [24], etc. Because humans can be regarded as experts in walking and push recovery, learning from human data may help in push recovery study and superior solution development. Therefore, for humanoid robots, finding push recovery control laws used by human beings may lead to an inspiration for better controller design and make up for the deficiency of engineering methods.

## 1.2 Project Goal and Contribution

Following our motivation, the project's primary objective is to find the most descriptive control law that people apply during push recovery within the chosen scope. In this project, we are interested in pushes that are intense and short in time such that the dynamics of human bodies are not changed. Before the start of the project, we have made a hypothesis of what control laws human beings may use. The hypothesis is that human beings use proportional-derivative (PD) control when simplifying human-body dynamics models to a one-dimensional point of the centre of mass. Relative information of this control law and the one-dimensional point mass model can be found in section 2.1. By designing and implementing a set of experiments for data collection and processing, we try to extract the centre of mass' motion trajectory, velocity and acceleration of a person during a push recovery trial and fit a model describing the underlying control law. Thus, if we input a set of position and velocity, we should get a reasonable acceleration prediction which is proportional to the force applied on the centre of mass using this control law.

To summarise, the contribution made in this project is a complete experimental design and implementation of a data-driven method including:

- A complete design of data collection experiment with prepared equipment.
- A complete data processing methodology of collected raw data for target dataset generation.
- Model fitting and evaluation based on regularised linear regression as well as statistical analysis. The models used are in the family of our hypothesised one.
- A set of concrete results is found within the designed project scope and proving our hypothesis. Human beings tend to use PD control for push recovery, and we also find that they have distinguishable sets of PD gains for different motion phases of a push recovery trial.

## 1.3 Organisation

This thesis consists of five chapters including this introductory chapter, and the rest of the thesis is composed in the following manner. In chapter 2, we start by introducing important concepts in push recovery study and some essential theoretical background on which our design is built. We also include our literature review of previous studies on push recovery in robotics in this chapter. Moreover, chapter 3 presents all methodology and designs we use in the project from data collection to data processing and data analysis. After that, we put all results, evaluations and the discussions on the outcomes we have of this project in chapter 4. Finally, we have chapter 5 to present the conclusions we have about the outcomes of our experiments, some critical analysis of our experiment designs and operations, and some suggestions on future work which worth exploring.

# Chapter 2

# Background

In this chapter, we provide the necessary background information related to this project. This information includes essential concepts and terminologies in human gait and push recovery study, recovery strategy and previous studies on push recovery which contain modelling, engineering controller designs and data-related methods.

## 2.1 Important Concepts

It is necessary to introduce some essential concepts and terminologies which play essential roles in walking, stability and push recovery, summarised by C. McGreavy [39]. This piece of work also inspires the layout of the subsequent section 2.3. Some concepts of the one-dimensional point mass model and a classic control law which form the foundation of this project are also introduced in the following subsections.

When a foot of a human is in contact with the ground, a support area is created on which the mass of the human can rest. Each of the feet has a supporting area, and by connecting the edges of all supporting areas via the shortest possible path, a **support polygon** can be given. This support polygon represents an area between the two legs where the **centre of mass (CoM)** should be projected on for the subject to stay balanced. The subject loses its stability and will fall over if the CoM moves outside of the support polygon. Maintaining this balance when faced with external forces is a goal of push recovery.

The ground evokes forces acting on the feet when contacts are made. These forces, known as **Ground Reaction Forces (GRF)**, have the same magnitude but opposite di-rection compared to the forces that contact feet apply to the ground. The force applied by a disturbance can be counteracted when applying correct GRFs, so it is crucial to regulate GRFs in push recoveries. Applying GRF in a moment arm around the CoM causes a torque around it, which is vital in order to reduce angular momentum. **Centre of pressure (CoP)**, on the other hand, is the point which we can sum up all pressure on a contact surface and form a force to act on.

Furthermore, the **Zero Moment Point (ZMP)** [51] is a point on the ground which guarantees no rotation around the horizontal axis happens when it is included in the foot support. When in situations where it coincides with the **Centroidal Moment Pivot** [41], no rotation around the CoM occurs, thus it becomes important as the CoM should be kept stable during push recovery.

Some information about the human gaits can also help because push recovery also contains standing state changes and making steps. The walking process contains cyclic gaits. According to the work of Kirtley [30], one gait cycle can be divided into two phases, stance and swing phases. During swing phase, the toe of one foot leaves and another contact between the same foot and the ground is made. Moreover, the stance phase starts at the initial contact of one foot and before the contralateral foot's toe-off. Between the extremes of a gait cycle, double support and single support can happen, representing the state of both feet being on the ground, and the state of only one foot is in contact with the ground respectively. In this project, we are interested in push recovery happening in a stationary state which can be regarded as continuous double support. An example of a complete gait cycle is shown in Figure 2.1.

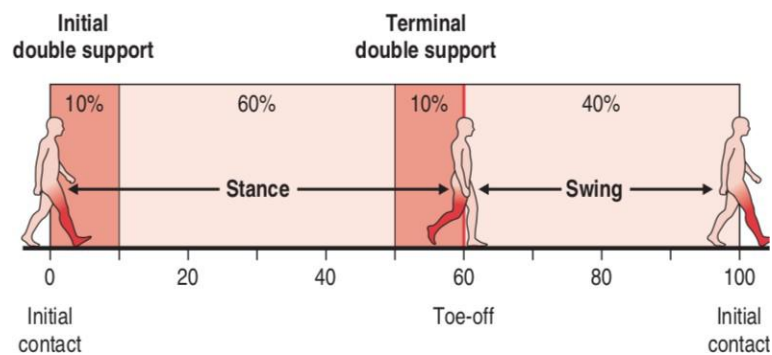

Figure 2.1: Diagram of a human gait cycle. *Source*: Figure 1.4, C. Kirtley *et al*. [30].

## 2.1.1 Proportional-Integral-Derivative Control

A Proportional-Integral-Derivative (PID) controller is an intuitive controller widely used in feedback systems. A basic single-input-single-output (SISO) feedback system can be described with Figure 2.2, where *r* is the reference input, *e* is the error signal, *u* is the control signal and *c* is the actual control output from the actuator and plant with $c_m$ being its measurement input back to the controller. PID controller is a simple but versatile control algorithm that computes the proportional, integral and derivative of the difference between the measured output control and the reference input to give a control signal to the actuator and plant. This output is therefore related to the defined PID coefficients and the input-output relationship is defined as:

$$u(t) = K_p e(t) + \int K_i e(t) dt + K_d \frac{de(t)}{dt} \qquad (2.1)$$

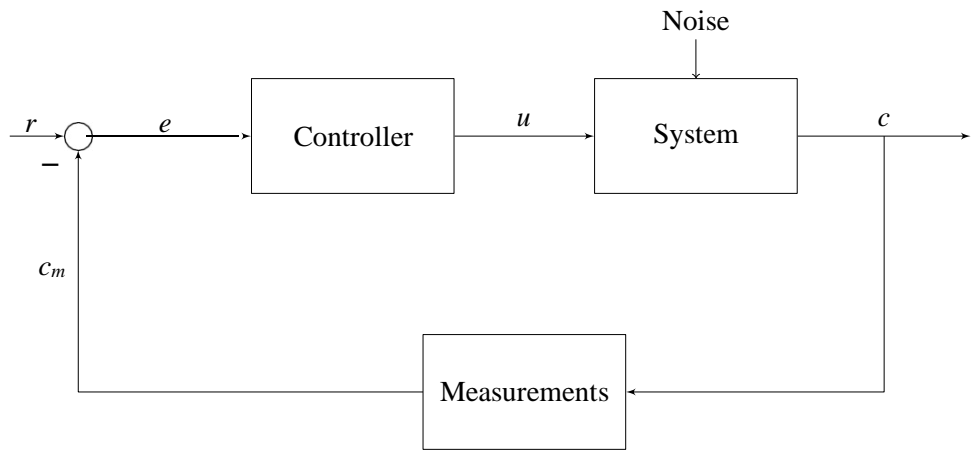

Figure 2.2: The block Diagram of a closed-loop SISO system

By leaving any of the terms, other control algorithms can be formed such as the proportional-derivative (PD) controller and proportional-integral (PI) controller. All three terms have different effects on the response's characteristics of the closed-loop system including steady-state error, rise/settle time and overshoot. When all three parameters are properly tuned, the system can reach optimal control which has the minimum cost function which can be the least energy for completing a task. Besides, PID control law and all its subsets have their terms in linear combinations.

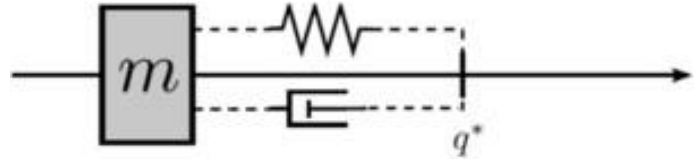

Figure 2.3: 1-D point mass dynamics, spring-damper.

### 2.1.2 One-dimensional Point Mass Dynamic Model

Here we introduce the concept of dynamics and a simple model which is used in the project's analysis part presented in section 3.3. The **dynamics** of a system describes how the control influences the system's state change. The simplest dynamic system is a one-dimensional (1-D) point mass model which contains no gravity and friction. Its **state** is then described by $x(t) = (q(t), \dot{q}(t))$ where $q(t)$ is its position and $\dot{q}(t)$ is its velocity. The control $u(t)$ is then the force applied on the point mass. Hence the system dynamics of this point mass is $\ddot{q}(t) = u(t)/m$ where $\ddot{q}(t)$ is the acceleration and $m$ is the mass. The 1-D point mass model represents a non-holonomic system, and a non-holonomic system contains differential constraints: $dim(u_t) < dim(x_t)$, meaning that not all degrees of freedom are directly controllable.

If we assume the point mass' current position is $q$, and the goal is to move it to $q*$. By adding feedback to the system, the control becomes

$$u = a(q^* - q) \tag{2.2}$$

If we add the velocity into the control law by having a desired velocity, the control is then updated to

$$u = a(q^* - q) + b(\dot{q}^* - \dot{q}) \tag{2.3}$$

We can see that the system dynamics formulations described by Equation 2.2 and Equation 2.3 are the same form of proportional (P) and PD control by referring to Equation 2.1. When under PD control, the point mass model becomes a spring damper shown in Figure 2.3. The subsets of PID control can all be applied to the 1-D point mass model following the same idea.

The 1-D point mass model can also describe push recovery process. The steady state of push recovery, standing still in a normal pose, can be defined as the CoM moves to the desired position $q^*$ within the support polygon area along the dimension defined by the push force vector. Its velocity and acceleration also reduced to 0 when reaching this steady-state point. This model can be more general by extending into a three-dimensional formulation.

## 2.2 Push Recovery Strategy

After presenting the essential concepts in the field, it is also necessary to introduce the strategies that people use for push recovery which have a significant influence on the recovery control policy. According to Hofmann's studies [19], three major push recovery strategies exist offering different classes of actions to deal with unexpected disturbances. The **ankle strategy** help restore balance using the torques generated at the ankle joint, while there is no bending of the hip. With the **hip strategy**, balance is restored by the combined use of ankle torque, and by the bending at the hip. It takes advantage of the fact that the CoM of a humanoid is usually located in the upper body. For push recovery, the CoM can be brought to rest over the CoP counteract the push force when applying the hip strategy. The last strategy is **stepping-out strategy**. This strategy is usually chosen when both ankle and hip strategies fail to neutralise the disturbance. To obtain the sufficient GRF to exert forces on the CoM, the last resort is to place a foot at a new position, by which the projected CoM may also move back in the support polygon. Foot repositioning strategies form the basis of a bulk of the research in push recovery, as this tends to involve recovery from pushes with larger magnitude. Diagrams showing these three strategies can be seen in Figure 2.4.

While it is easy to distinguish and apply different strategies to humanoid robots, it is not the same for human beings because when applying different strategies, e.g., ankle strategy. A person's hip cannot be isolated and kept complete rigid when doing ankle strategy compared with robots, so as other human joints. In our study, we define three basic and two combined recover strategies based on the observation we have during the data collection phase of the experiments, presented in section 3.1. These five strategies inherit the features of the three recovery strategies above while the natural state of human beings is also put into consideration. In further stages of our experiments presented in subsection 3.2.3, we use these recovery strategies to split our dataset into

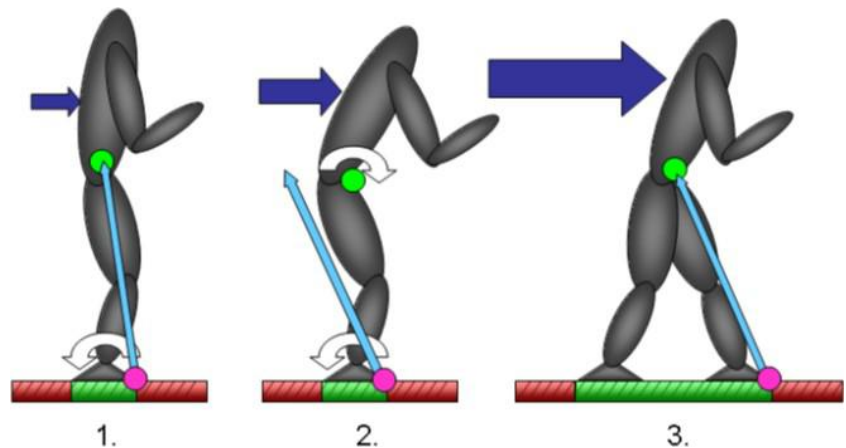

Figure 2.4: The three basic recovery strategies. In the figure, green dots represent the CoM, the magenta dot represents the CoP, and the cyan arrows represent the GRF. 1. Ankle Strategy 2. Hip Strategy 3. Stepping-out Strategy. *Source*: B. Stephens, 2007 [45]

different categories for analysis.

- **Ankle Strategy**: Essentially the same as the one in the literature study while a little natural bending of the hip and knee is acceptable and also taken into account.

- **Tiptoe (Toe) Strategy**: Corresponds to the hip strategy in the literature study. The recovery is accomplished by the combined motion of lifting heels to stand on tiptoes and usually combined with hip and knee bending.

- **Toe-to-step Strategy**: Recovery begins with tiptoe strategy, but the strategy is then switched to making a step during the process because of the judgement that the push is too intensive for tiptoe strategy to compensate.

- **One-step Strategy**: The recovery is accomplished by making one step ahead.

- **Multi-step Strategy**: The recovery is accomplished by making two or more steps ahead. In our project scope, we have only observed a two-step strategy.

## 2.3 Previous Studies in the Field

We summarise some previous studies with both engineering-based methods and data-driven methods in this section. The information presented in this section helps us in finding the motivation and setting up the project goal. To be specific, critical analysis of the methodology of some studies with data-driven methods significantly aids to our experimental design. This section is mainly based on the author's IPP report of this project [47] and the other source [39] already mentioned in section 2.1, but with more literature materials, insights and analysis of the contents included.

### 2.3.1 Model Simplification and Traditional Approaches

Researches on walking and push recovery with engineering-based methods use simplified models to represent the dynamics of a system such as a humanoid robot. Model-based controllers tend to have problems with full-body dynamics models because these models are difficult to construct in real-world scenarios and they are computationally expensive with a high probability of being inaccurate. The Linear Inverted Pendulum Model (LIPM) [27] describes the body dynamics of a simple pendulum with a point mass at its summit. The mass of the pendulum represents the upper body of a bipedal robot and is linearly constrained by an extensible leg with no mass. The author's later work extends the model's dynamics from the sagittal plane to 3D cases, introducing the 3DLIPM [26]. The LIPM has been a useful models but do not have a full generalisation in humanoid robots since its fundamental assumption is the foot movement being instantaneous without considering angular momentum. Further variations of the LIPM have also been introduced such as the Angular Momentum inducting inverted Pendulum Model (AMPM) [31] and the LIPM with flywheels [42]. LIPM variations have been used in push recovery controllers and may still contribute in future studies.

The deconstruction of the motion equations of the LIPM models derives a class of step alteration push recovery solutions. Takenaka *et al.* introduced the posited concept of the Divergent Component of Motion (DCM) [46] which uses the unstable aspects of LIPM motion equations to produce trajectories for the model's CoM and feet. A DCM-typed algorithm has been developed to predict the Instantaneous Capture Point (ICP), a point on the ground which must be included in the base of support to bring the robot to a complete stop given a known state. The robot gets into a capture state when the capture point is reached. In this state, the robot accomplishes energy equilibrium thus becomes stable and will not fall over. The ICP algorithm is derived based on the LIPM, and the capture point's position is calculated using the system's two eigenvectors depending on the velocity and height of CoM. According to the assumption of the LIPM that CoM has a constant height, there is a unique capture point for every state given specific push force. When using the LIPM with flywheel, capture points can also be derived for bodies with angular momentum. This ICP paper gives a milestone for push recovery and balance research because eigenvector stability ensures optimal stepping location theoretically and the computed ICP is a desirable target stepping location. However, because of the LIPM's instantaneous movement assumption, the algorithm has unsolved implementation issues on real-world robots. Nevertheless, it is still a powerful tool, and its current modifications have been used in walking controllers [14][12][32]. Work of Englsberger *et al*. [13] introduced the DCM in another form that outputs a point in 3-D space where the CoM must move to in order to become balanced. This work is an example of a generalised walking controller which is also useful for push recovery. Although its calculation may not guarantee feasible solutions in hardware implementation, its concept has been extended and used in the DARPA robotics challenge [22].

Similar to the ICP algorithm, an online predictive controller has been done using the LIPM and robust extended model predictive control to control foot placement [8]. This control method takes a biped's actual CoM state feedback to compute future foot placement and re-plans ZMP references and foot trajectory analytically. Therefore, the control method enables a bipedal robot to have dynamic and reactive walking which is robust against external force disturbances. Besides, work has also been done on the comparison between the LIPM and the inverted pendulum model(IPM) in balance recovery [34]. By resolving the analytic issues of the IPM with acceptable approximation, the foot placement estimator is extended by analytically derived predictive formula. According to the paper, the IPM includes step transition and have smaller step distance, better leg reachability, lower knee torque and higher disturbance rejection in balance recovery control compared to the LIPM. These results may inspire passive dynamic walking(PWD) principles to be extended in powered bipeds, and filling the gap between the ZMP and PDW communities.

Beside the DCM, other types of online walking planning methods also exist. Preview controller tracking a ZMP trajectory introduced by Kajita *et al*. [25] is of another type. By solving the CoM's divergence problem, the controller is able to perform well against strong pushes by quick reaction to rapid changes in ZMP direction. It takes only speed and direction as input and can produce new ZMP-CoM trajectory online.

However, although this new trajectory is a desired property, the output ZMP motions may go beyond the reachability of the robotic platform with this controller implemented since the controller cannot be constrained.

In addition, if a walking trajectory computed offline encounters disturbances, the system needs to reject pushes while the walking controller needs to keep track of the original trajectory in order not to plan again. The work done by Komura *et al.* [31] introduced a feedback controller augmenting a feed-forward walking controller doing stepping plans. When a walking trajectory is computed, the performance of the robot following the trajectory is tracked by the feedback controller. When the robot encounters a disturbance such as a push, the feedback controller then tries to find a new stepping position that will help the robot recover to balance while the profile is kept as close to the original position and velocity calculated by the feed-forward controller as possible.

Finally, optimisation based stepping planners tend to plan according to footstep timings with a set of swing time. A step which requires modification in the face of a distance will make its plan be abandoned. Several related methods have been introduced in this category. For example, the work of Griffin *et al.* [18] uses expected swing time in an ICP based control algorithm to adjust for external disturbances in an attempt to solve the timing issue for large ICP errors. Besides, the work of Kamioka *et al.* [28], in which a method of push recovery with step adjustment while bypassing the time issue, also belongs to optimisation based step planing. Learning methods can also be used in step timing issues. For example, an online function approximation was used by Vijayakumar *et al.* [50] to learn a push recovery control policy during live walking. Instantaneous movement assumption issues in ICP and step timing issues were both solved by this learning methods. However, the required amount of data went huge due to the continuous push direction, magnitude and location, making the feasibility of this option questionable.

Until now, various traditional methods have been proposed and adopted in simulation and/or real-world implementation. These methods considerably advanced the field of walking and push recovery study, and can still be developed and inspire discoveries.

### 2.3.2 Human-inspired Studies

Explorations in human-inspired studies also exist in walking/push recovery studies. For example, some experiments have been done to study the relation between push recovery strategy selection and the push force intensity [29]. In the experiment, a boxing glove with a force sensor placed on it is used to apply pushes and measure the intensity and direction of the push force. Besides, a body-mounted inertial measurement unit (IMU) is used to measure the human body's inertia and acceleration. This experiment is composed by applying different pushes to the test subject from his front, back and side direction. Data collected from the 78 trails in the experiment give a useful result being able to classify whether the subject is provoked to take a step based on the applied push force and the final accuracy reaches 79%. However, because the push is applied manually through a boxing glove, information gathered by the force sensor can be inaccurate and manually applied pushes can also be unsatisfactory in precision. Also, since only one IMU is used to take measurements from the human body, information of the support polygon, GRF as well as ZMP is missed although the IMU can measure the acceleration with direction applied to the CoM accurately. Moreover, the experiment only involves one participant for data collection, which limits the generalisation of the result. These aspects limit the experiment from building a more precise strategy selection of human push recovery. Information about the participant's prior knowledge of an incoming push is also missing, and whether the participant has predictive movement or not may profoundly influence the push recovery process.

Experiments have also been done on eight human subjects to verify the ICP algorithm [1]. In their attempt, the constant height assumption of a LIPM is removed and replaced with the assumption that the body is in a free fall state during the swing phase when making a step. It also assumes the forward velocity is constant after the push. The experiment's outcomes show that the modified algorithm is capable of predicting the foot placement for a push recovery trial to some extent but it still has some limitations, and possibly some factors have not been taken into consideration during its development. These factors include upper body inertia, energetic cost and the LIPM based models' relatively low generalisation quality mentioned in subsection 2.3.1, etc. The finding of this experiment is shown in Figure 2.5 suggesting a person's capture region, a small region around the ICP, shifts from the predicted ICP capture region in the forward direction. Nevertheless, the ICP is still important because of its theoretical value, even if the average distance between real foot placement points and corresponding predicted ICP is $11.4 \pm 5.8cm$. This distance decreases when restrictions are added to the subject. Because full-body dynamics models can be used in simulations with human data, data-driven methods can involve the factors causing inaccuracy to the ICP related algorithms thus lead to more precise predictions.

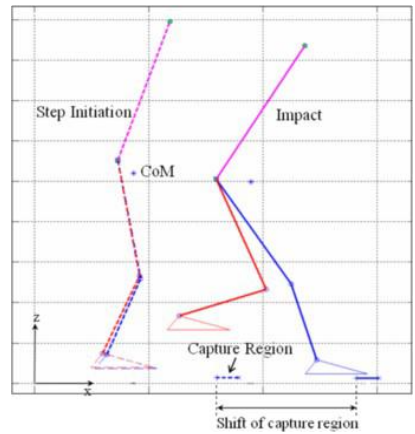

Figure 2.5: ICP test with human data on LIPM models. *Source*: Z. Aftab *et al*. [1].

There are also other kinds of studies in human motion. Kulic *et al*. [33] introduced a method to segment continuous human kinematic data into hierarchical motion primitives. Their study focused primarily on the segmentation performance, which seemed to have a low false positive and negative rate during segmentation of motion into distinct categories. However, the method's real-world practicability could not be accessed since no implementation on robotic platforms were done to generate motions. Nevertheless, this human motion segmentation methodology has been applied in computer graphics and animation for general motion generation [20] and specifically, push recovery motion generation [55]. Yin *et al*. [55] adapted motion capture data of human push recovery to animated characters by finding the nearest neighbour in the experimental data profile using a predefined input push force to generate specific responses. Holden *et al*. [20] implemented a deep learning approach to infer human motion based on supervised learning of human motion data, especially for future natural movement prediction. These simulation studies also contain potential in future robotics applications.

For real-world application, biomechanical study of human locomotion has advanced human-inspired robotic studies. For example, the under-actuation phenomenon during standing balance and walking revealed in [53] and common observation of human balance recovery with foot titling inspires the study given by Li *et al*. [35]. Combining CoM variable impedance control, nonlinear virtual mechanical stoppers as well as torso attitude and posture control, the proposed control framework effectively reproduces human-like balance recovery behaviour with active foot tilting in sagittal scenario. This behaviour is also successfully replicated on a real humanoid robot, which gives a promising foundation for future studies.

In addition, other studies translating human data to implement on real robots also exist. Imitation learning methods exploit human data to generate walking, grasping, and other motion policies for robotic hardware [43]. It is possible that good push recovery motions being translated to biped robots on the base of human data learnt by imitation. On the other hand, neural science studies also show the possibility of improving motion generation in robotics. Central Pattern Generator (CPG) proposed by Brown [6] and applied to locomotive robots by A. J. Ijspeert [23], for example, enabling the controller to generate a walking profile with only a few parameters to tune, which can also help in push recovery controller development. However, until the start of this project, we have not found relevant studies in directly finding the push recovery control law applied by humans. Therefore, this area remains to be explored, and successful findings should be able to contribute in the field and inspire future push recovery controller design.

# Chapter 3

# Methodology and Experiment Implementation

In this project, we use a data-driven method to find a model describing the control law that people use for push recovery. To be more precise, we intend to find a control law describing the relationships among the CoM's position, velocity and forces that the subject decides to be applied in order to compensate the push. By extracting the CoM using simulation on collected human data, we simplify the complicated human dynamics model to a 1-D point mass model with its states defined by the position and velocity, and its control is the force applied on it. Therefore, the control law should be able to compute an output acceleration (proportional to force) given an input pair of position and velocity. The analysed dimension is the one defined by the push applied to the model. We hypothesise that human beings use PD control for push recovery when the full human dynamics is simplified to this 1-D point mass model.

The experiment implementation of this project can be divided into three main phases: Data collection, data processing and data analysis. Therefore, three sections are set corresponding to these main phases and are presented in sequential order.

## 3.1  Data Collection

This project is based on a data-driven method where human data become necessary. Therefore, some participants are needed for data collection. In this project, we had both male and female participants aged in their twenties and thirties, with their height and weight picked randomly. All participants were healthy individuals with no disabil-ities. In addition, all participants fitted the requirement of being able to walk naturally and recover from a push with its intensity close to common real-life situations.

There were 10 participants taking part in the experiments in total, and all of them were well-informed by the provided experiment information sheet and the researchers' oral explanations. All related document templates including the information sheet and consent forms can be found in Appendix B. Consequently, all participants accepted the listed conditions and volunteered to join this experiment. To add on, this data collection phase is associated with C. Mcgreavy [1] from EPSRC, University of Edinburgh.

### 3.1.1 Musculoskeletal Models

Simulation is necessary to compute the CoM's state properties. In this project, we decided to use the software OpenSim [10] to simulate motion and compute the CoM's position trajectory throughout a push recovery trial. Because lower limbs and torso which contribute the most in push recovery have a significant effect on the CoM position, we used a pre-existing OpenSim human musculoskeletal model, the Gait-2392 model [2][3][11][54] which has 92 muscles and 23 degrees of freedom. This model monitors full body kinematics without upper limbs and is often used in lower limb motion simulation analysis.

Other models available in this stage of our project are usually specific extreme models such as the Human Neck Model [48], Upper Extremity model [21] and Lower Extremity model [11] which cannot be used to simulate full body dynamics. On the other hand, some other similar full-dynamics models also exist such as the Lower Limb Model 2010 [4][11][5]. This model contains extra utilities including ellipsoidal wrapping surface which makes its dynamic analysis much slower than that of Gait-2392. Putting all these aspects into consideration, we decided to use the Gait-2392 in the end.

To fit motion capture data to this model, we designed a maker set based on the APO model from the work of Gordon *et al.* [17] for exoskeleton performance experiments. Since no exoskeleton is attached in our experiment and the mark sets are almost identical, we named the model in this project 'no-APO model'. The generic no-APO model with the marker set on can be seen in Figure 3.1 which is used to scale every partici-pant's model in the data processing stage presented in subsection 3.2.2.

---

[1] https://www.edinburgh-robotics.org/students/christopher-mcgreavy

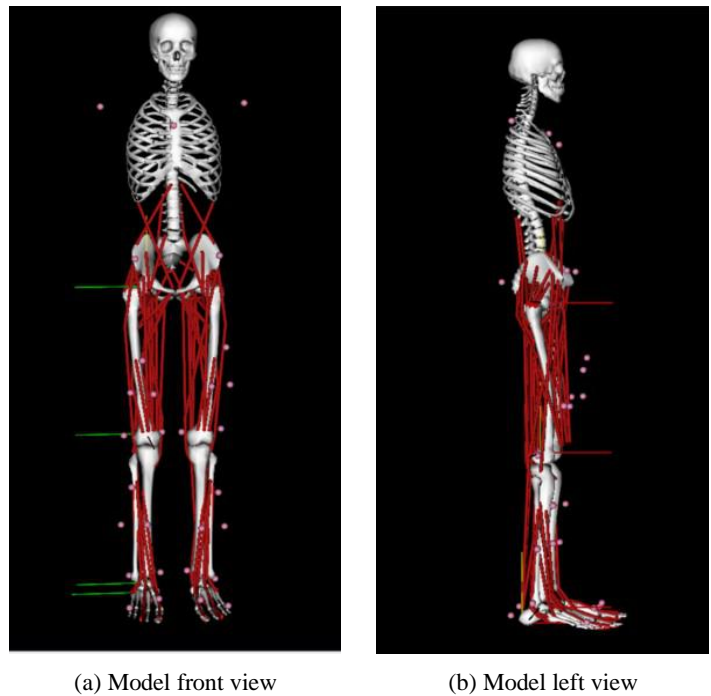

(a) Model front view   (b) Model left view

Figure 3.1: The generic model with the marker set

### 3.1.2 Experiment Design

Learnt from some previous studies with data method, we come up with the experiment design of this project. Data are collected for each participant standing on a treadmill and experiencing push recovery. Unlike [29], a push is generated by the treadmill's stopping with different initial speeds such that the 'push force' is always fixed in the horizontal direction and can be regarded as applied directly on the CoM. Ground reaction forces (GRF) and speed data are collected with this six-axis, split belt instrumented treadmill (Motekforce Link, Amsterdam, Holland). Also, reflective markers are attached to all participants in order to track their movements accurately, and motion data of a push recovery trial are recorded with a six-camera motion capture system (VICON, Oxford, UK).

In this project, a marker set consists of 33 Cleveland markers is used and 8 of which are used only for scaling the dynamic no-APO model presented in subsection 3.1.1. An example of reflective the marker set-up for participants is shown in Figure 3.2. Four plates with markers attached are used on thighs and shanks, indicating these parts of the lower limbs are rigid bodies. Other individual markers are placed on body landmarks and joints' centres of rotation such as knee and ankle markers. In order to make marker positions more persistent, the relative motion of clothing should be reduced, so all participants are asked to wear shorts and slim-fit T-shirts or a set of gym wears as well as relatively thin-sole footwear in advance. After marker attachment, the participant needs to get on the treadmill and have the same T-pose shown in Figure 3.2a for static pose capture. This static capture will be used for further model scaling.

Before the start of data collection, each participant is connected with a safety harness and asked to stand on the right edge (from the front view) of the treadmill in their natural pose with comfortable foot spacing. This spacing should be roughly the same as that during the static T-pose recording. When in this pose, we changed the distance between the two projected foot placement reference areas such that the participant can return and prepare for each trial with the same starting position and foot spacing. The complete experimental set-up can be seen in Figure 3.3. After setting up, a few (usually fewer than 5) practice trials are carried out without recording to help the participant get familiar with the experiment procedure and jerks caused by the treadmill's motor at the beginning of each trial. Before the start, all participants are asked to try to use a recovery strategy with minimum control; i.e. if using other softer strategies mentioned in section 2.2 can deal with the current push, try not to make a step, but do switch strategies if necessary. Experimental procedure for formal trials can be summarised into the following major steps. More information about this experimental procedure set-up can be found in Appendix A.

1. Let the participant stand on the projected starting points.
2. Start the treadmill with one speed level and the recording starts automatically after the speed reaches the desired speed of the current trial.
3. Stop the treadmill suddenly to generate a 'push' when the distance moved goes over half of the treadmill's length. The treadmill's deceleration is kept at its maximum, $3m/s^2$.
4. Wait for 3 seconds after stopping the treadmill and automatically stop the recording, during this time the participant should have fully recovered from the 'push'.
5. Let the participant go to the start and prepare for the next trial.

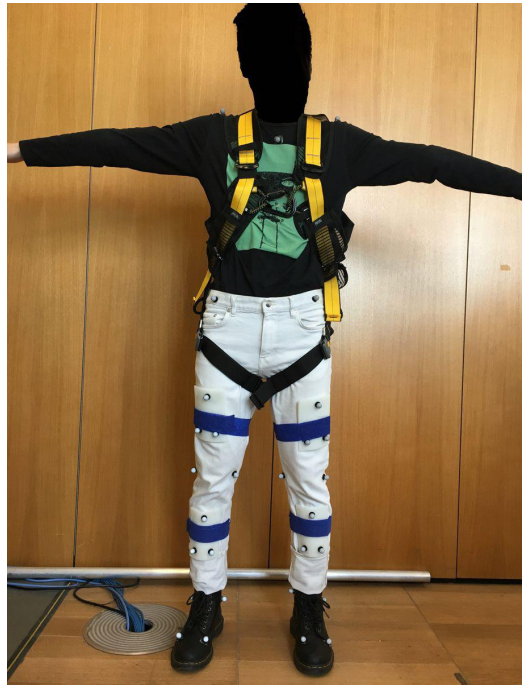

(a) Front view

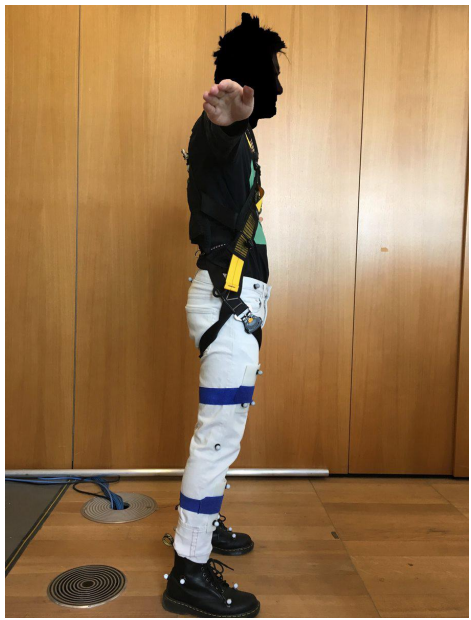

(b) Side view

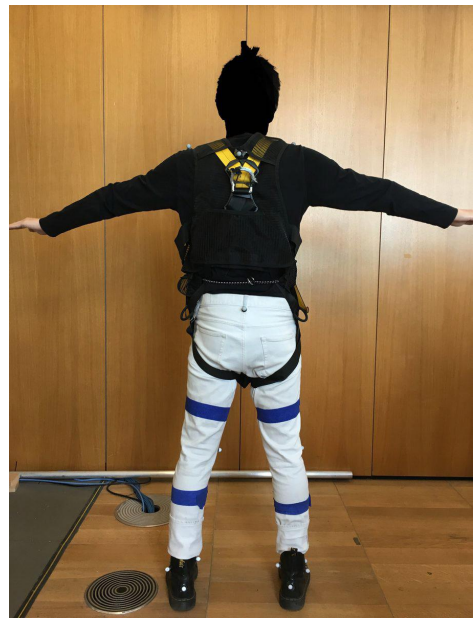

(c) Back view

Figure 3.2: A marker set-up example, with the harness vest on

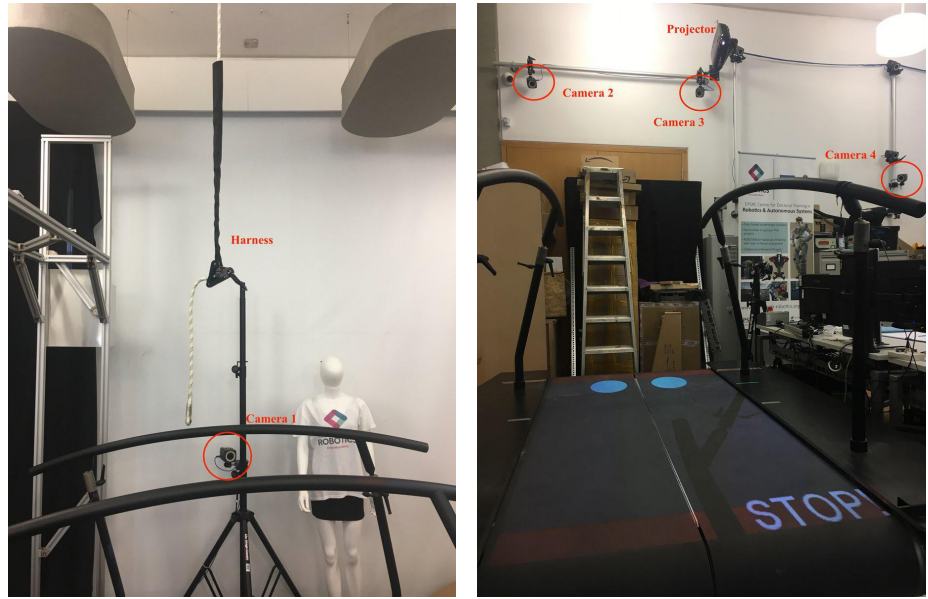

(a) Front view: Harness and Camera 1  (b) Left view: Projector and Camera 2 - 4

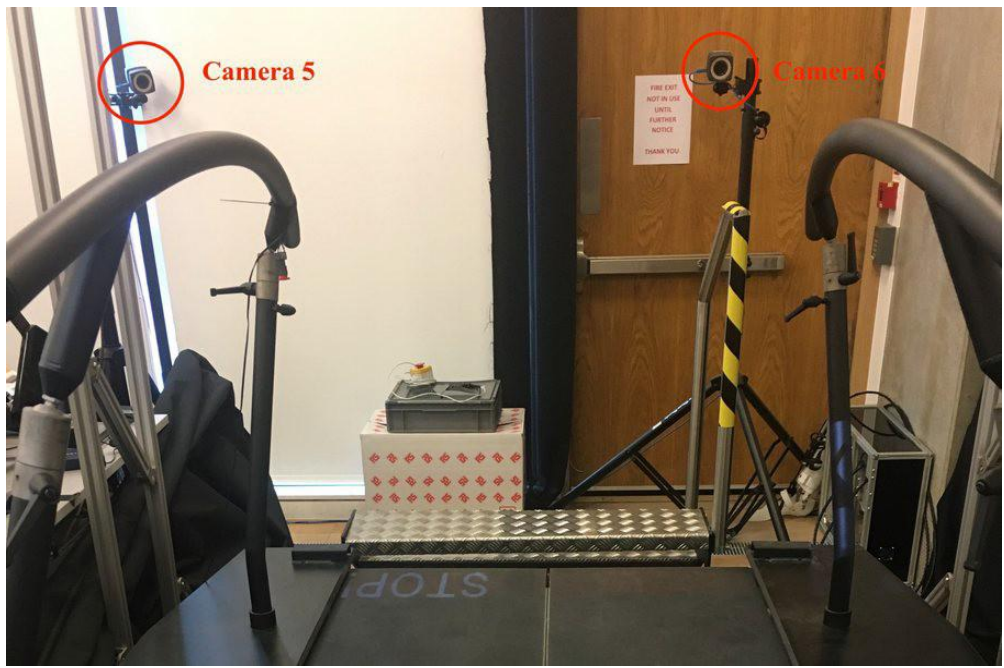

(c) Right view: Camera 5 & 6

Figure 3.3: Experiment set-up: Treadmill, projector, harness and camera set

We design 10 different speed levels with 5 trials in each level, for every participant. The speed range is from 0.6*m/s* to 0.825*m/s*, with a step difference of 0.025*m/s*. Hence, there are 50 trials in 10 conditions, 5 trial per condition, prepared for each subject.

When testing the experiment design, we find that people may unconsciously try to adapt to the procedure by predicting the incoming push and react in advance. To soften this issue, we evenly split the 50 trials into two parts. In the first 25 trials, every participant receives a countdown from the researcher before the treadmill starts to move while in the later 25 trials the researcher only mentions that the trial is going to begin soon, and randomly waits for several seconds. By having this unexpected trial starting scheme, we intend to reduce the effect of participants getting adaptive. Also, in scenarios without sufficient information, we also expect that some changes may be applied to in their push recovery behaviours.

To make the whole process more automated, we implement a Lua script in the Motek D-Flow software to uniformly random pick an initial speed of the treadmill and trigger the recording command when a trial starts and ends. For time synchronisation, the Lua script implemented in the D-Flow software can send a command to a relay box which triggers the VICON system, Motek treadmill to start data recording simultaneously, with the internal delays also taken into account. The motion data, GRF and speed data were captured at 100 and 600 Hz, with the VICON system and the Motek treadmill respectively.

### 3.1.3 Post-processing of Raw Data

Because of the camera placement, the 100Hz record frequency and marker position instability introduced by a participant's motion during a trial, a marker can be missing, dropping its label, switching its label with other markers, or having their relative positions stretched during one trial. Some demonstrations of these situations can be seen in Figure 3.4.

Therefore, some post-processing of the captured motion data is necessary in order to have the model of a participant consistent in every trial. The VICON Nexus software provides the several gap filling methods, with their descriptions in the user guide [49] presented as follow:

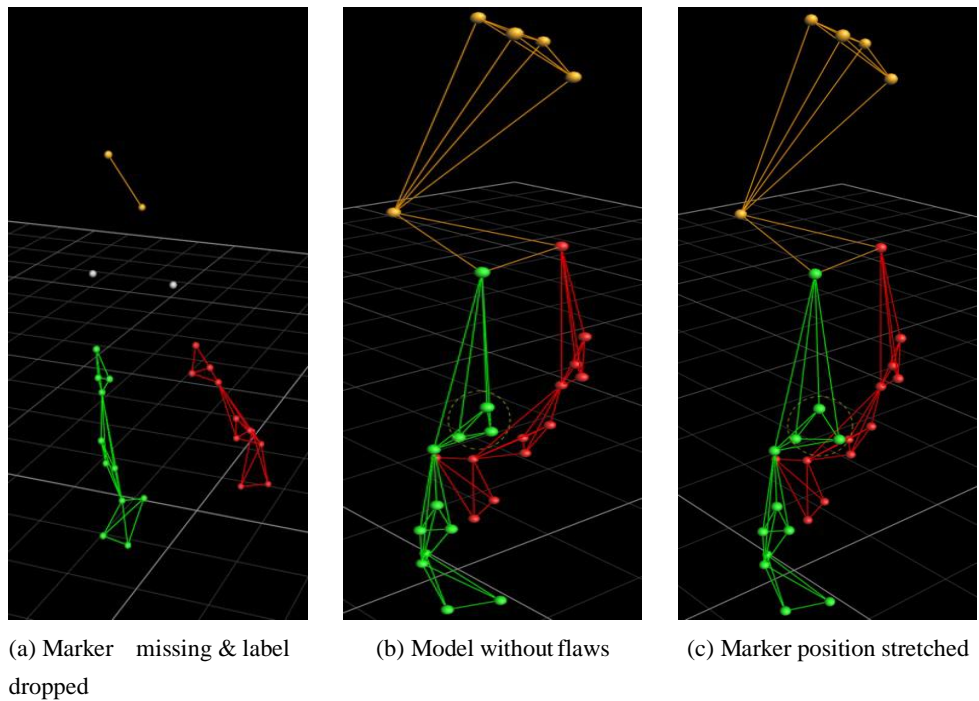

(a) Marker missing & label dropped

(b) Model without flaws

(c) Marker position stretched

Figure 3.4: Example of issues in VICON recordings which requires post-processing

- **Spline Fill**: Perform cubic spline interpolation operations to fill selected gaps.
- **Pattern Fill**: Fill the selected gap using the shape of another trajectory without any gaps.
- **Rigid Body Fill**: Fill gaps based on rigid or semi-rigid relationships among selected markers.
- **Kinematic Fill**: Fill gaps using information about the connection of markers to segment in the labelling skeleton template. Kinematic Fit pipeline operation required to make this option available.
- **Cyclic Fill**: Use the missing marker's pattern from earlier or later gait cycles to fill gaps, for naturally repetitive motions, e.g. walking on a treadmill.

According to the descriptions of gap filling tools, our gap filling scheme can be summarised into the following statements. The first of which is to use spline fill for all gaps whose length is below 20 frames. Pattern fill is used when the missing marker has another marker with the same or a highly similar trajectory pattern, e.g. the clavicle and C7 (on the top of the back) markers. For rigid body fill, we use it based on some rigid body assumptions of corresponding areas. These rigid bodies are the set of markers on feet, including ankle ones; The two hip marks and the sacral marker. The reason of including markers on the joints is that all joint markers are placed on roughly the centres of rotation of corresponding joints, which can be regarded as fix points. Hence we can also include them into the rigid bodies formed by other nearby markers. Moreover, for some part of a trial, e.g. before the push recovery starts, we also assume that the four top markers (clavicle, lift and right acromium, C7) form a rigid body since usually, they have only little relative motions during that period of time. Finally, if all the above operations are not available, we turn to cyclic fill because some parts of a push recovery trial can be recognised as cyclic movements by the software and it has achieved satisfactory results throughout the experiments. Because we do not have related pipeline operation prior to the gap fill, the kinematic fill is not used in our experiments.

On the other hand, in order to deal with stretched marker position, we undo the marker's label in the corresponding frames and use gap filling to compute a more reasonable trajectory of that marker. When a markers label is found dropped in a trial, we have manual re-labelling done to it. In addition, markers' trajectories, especially the four up-most markers, are traced through the trial. The trajectory of any marker switching its label with other markers will become inconsistent at some frames and we then do the labelling again from that node to fix the issue. Sometimes the system never captures a marker again when the marker becomes missing. If this situation happens, we cannot do gap filling because the software cannot specify gaps under this situation. Therefore, we trim the trial having this issue if possible, and those issues cannot be solved here is handled in subsection 3.2.1.

After all these operations the recorded motion file should no longer have potential issues introducing significant error in the data processing stage. Although markers may have fluctuations because of small movements of clothing and measurement noise of cameras, most of the fluctuations can be dealt with at the beginning of the data processing phase presented in subsection 3.2.1. However, sometimes fluctuations can cause a substantial shift to a marker's position which is visually apparent. If this case happens, using the same strategy as stretched marker position mentioned above can deal with it. After raw motion data post-processing, we export all recordings into C3D files (.c3d) including both trajectory data and static models, to process in the next stage.

## 3.2 Data Processing

After data collection, we have the raw data recorded in three kinds of files including C3D files for a participant's static pose model and marker trajectories of each trial, and text files (.txt files) for speed and forces respectively. We have a toolbox with OpenSim APIs to analyse our data and compute CoM's properties (position, velocity, acceleration) and generate MATLAB formatted data (.mat) files, presented in subsection 3.2.3. However, since this toolbox cannot directly take these raw data files, we need to process our raw data with some adjustments if necessary, and generate files in the target format. We also have a software list for the project which can be found in Appendix A.

### 3.2.1 Input File Generation

The raw data files we have cannot be directly used for data processing with OpenSim. To be able to do that, three kinds of files are needed including marker trajectory recording (.trc) files, GRF recording (.mot) files and a scaled OpenSim model (.osim) file of the corresponding participant. In this project, a MATLAB toolbox, NOtoMNS, developed to process motion data developed by A. Mantoan *et al*. [36] is used to accomplish this job. While instead of using the original version, we changed the source code to a private version by adding automation functionality to process our batched data. There are two major tasks to do at this stage, dynamic and static elaborations. The flowchart of the dynamic elaboration is shown in Figure 3.5. A 6 Hz low-pass filter is used for data filtering which is done for both force and trajectory recordings to deal with noises and get smooth data. After dynamic elaboration, we will get all trajectory and GRF data in the desired format. Static elaborations are executed next to generate marker description files in '.trc' format with joint centres' positions calculated for model scaling in the next stage, presented in subsection 3.2.2. We have all trajectory files checked after generation because some files contain 'NaN' values indicating missing data which can cause errors in further processing. The reason of having this situation is that during raw data post-processing, some markers never become visible again to the camera set after missing, where we cannot do gap filling, as described in subsection 3.1.3.

### 3.2.2 Model Scaling and Adjustment

Having all static pose files generated in the desired format, it is now possible to scale OpenSim models for every participant. To do model scaling, we load the generic model

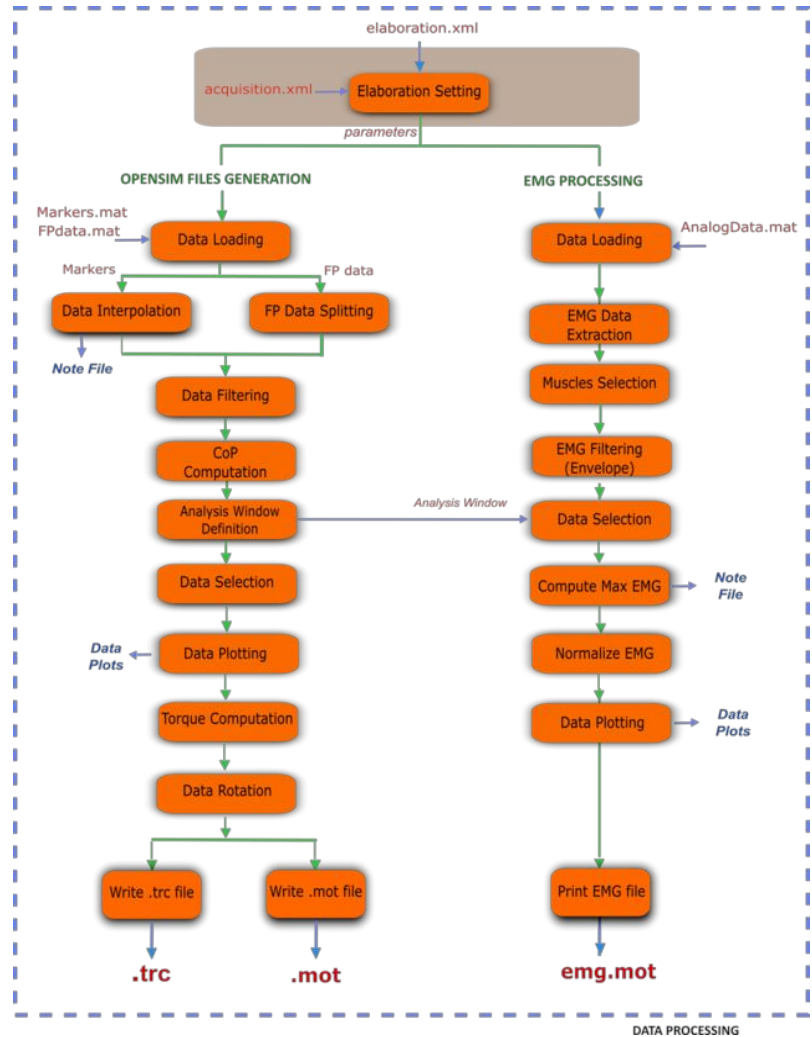

Figure 3.5: Dynamic elaboration flowchart of NOtoMNS toolbox.    *Source*: http://rehabenggroup.github.io/MOtoNMS/manual/dataProcessing.html

'no-APO' presented in subsection 3.1.1 in OpenSim software and use the software's scale tool. By having a prepared description file and the marker description (.trc) file generated in static elaboration, it is possible to run automatic model scaling and generate a scaled model for each subject. To note that, we also include the mass of each subject in the scale tool for future stages. Because the scaled model comes is based on the recorded static pose recording only, it is necessary to examine how this model generalises to all dynamic trajectories of recorded trials. Therefore, the inverse kinematics tool is used for this evaluation. Here we also prepared a description file for the inverse kinematics evaluation. By loading both this file and a selected trajectory recording (.trc) file, we are able to run the analysis and see errors on corresponding marker pairs between our scaled model and the recording. If the global maximum error is less than 4*cm* and the accumulated RMS error is less than 2*cm*, then this scaled model passes the evaluation and can be used in future steps, according to the OpenSim documentation [40]. This evaluation can only be done manually, but most motion recordings of a participant are loaded and analysed to make sure the scaled model passes the evaluation in general.

For some participants, the static and dynamic marker trajectories do not match up because some makers' related positions were changed during data collection. This mismatch can lead to the pose of the scaled model stretched by the software during the inverse kinematics analysis to reduce the error as much as possible since the software cannot modify the markers' related positions according to our set-ups. An example of a stretched model can be seen in Figure 3.6. We can see that the model stands on its tip-toe as a normal pose which does not match the reality. Hence some adjustments are needed to solve this kind of problems. To do this, we load both the static and a dynamic trajectory. By comparing the two trajectories, we can see some related position differences. These differences are then referred, and we change the generic model's marker position to make a more suitable version for the participant. The modified generic model is scaled and re-evaluated following the procedure presented above. We keep the new scaled model if it passes the evaluation. Otherwise, we need to go through the same procedure again until the model can pass the evaluation.

After finishing all adjustments and evaluations, we save all scaled models satisfying the requirement for every participant and pack them with other related files to use in the next stage of data processing.

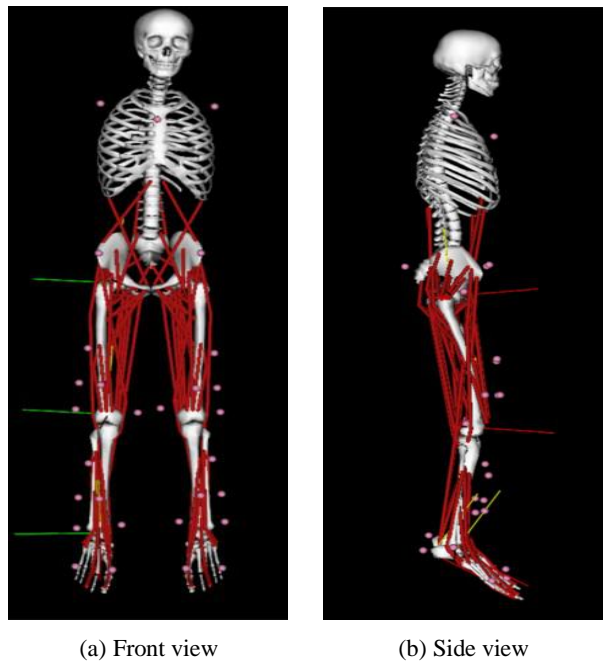

(a) Front view      (b) Side view

Figure 3.6: Example stretched model.

The model always stands on tip-toe and does not look in the front direction

### 3.2.3 OpenSim Analysis, Dataset Generation and Post-processing

Having obtained scaled models, trajectory and GRF data in the desired format, we can now move to data processing. Here we use another toolbox developed by D. Gordon [17] which uses OpenSim APIs to do automatic analysis and output file generation. The analysis tools we use in this project and their functionalities are as follow:

- **Inverse Kinematics (IK) Tool** is used to calculates joint angles based on marker trajectories.

- **Residual Reduction Algorithm (RRA) Tool** takes dynamic inconsistency between the musculoskeletal model and the measurement data [10]. It is used to generate dynamically consistent joint angles and a corrected model file from the joint angles and GRFs.

- **Inverse Dynamics (ID) Tool** is applied to calculate joint torques using the RRA-corrected joint angles and GRFs.

- **Analysis Tool** is used to calculate the position and velocity trajectory of the model's CoM given the RRA-corrected kinematics.

After all tools' analysis completes, we can use the results in the analysis tool and generate MATLAB formatted datasets of each participant. However, when running the toolbox, some errors occur during the analysis of the RRA tool and the analysis process is aborted. The error indicates that the GRF data cannot match with the trajectory data. When loading the GRF data and corresponding trajectory data, we find that the duration of the pair of files do not match and the GRF data appear to have longer recordings. The Lua script we implemented in D-flow software for automation described in subsection 3.1.2 can be the cause of this issue. Fortunately, cutting the redundant data based on time steps can solve this issue. Therefore, we set a time offset for each subject and run the NOtoMNS toolbox again to generate another set of files. The RRA analysis does not have any error with the new data files, and when we load them in the OpenSim software, we can see the motion and GRF forces are visually matched up.

Because the GRF still does not completely match up with trajectories, and our experiment include unexpected torque caused by the treadmill, the RRA error usually goes above the reference range [40] which will influence the results of the ID tool. Nevertheless, this situation does not affect our data because the data contained in our target output file are essentially computed by the IK tool which does not rely on the dynamics analysis. Hence, we can ignore the results. However, the RRA error is still kept below 100 because it also suggests how well do the GRF and trajectory data match up. The overall data processing pipeline of this project, starting from subsection 3.2.1 till this subsection, can be summarised by the block diagram shown in Figure 3.7.

Having all processed data in the desired format, we enter the final stage of data processing which is extracting desired data and trim all trials to get only the push recovery part. Calculating the force vector is equivalent to calculate the CoM acceleration because during the push recovery force applied on the CoM is $F = ma$ where $m$ is the mass in kilograms, and $a$ is the acceleration in $m/s^2$, ignoring the air resistance. We have considered one way to use the recorded GRF data from the treadmill, according to B. Stephens' work [45], the acceleration of the CoM and GRFs have a relation that GRFs can be assumed to be directly analogous to forces applied to the CoM such that *Acceleration = GRF/Mass*. However, because of the redundant data in the recorded GRF files, and the fact we only shift the GRF data to make them visually match up with the motion of a trial, accelerations calculated in this way can be unreliable.

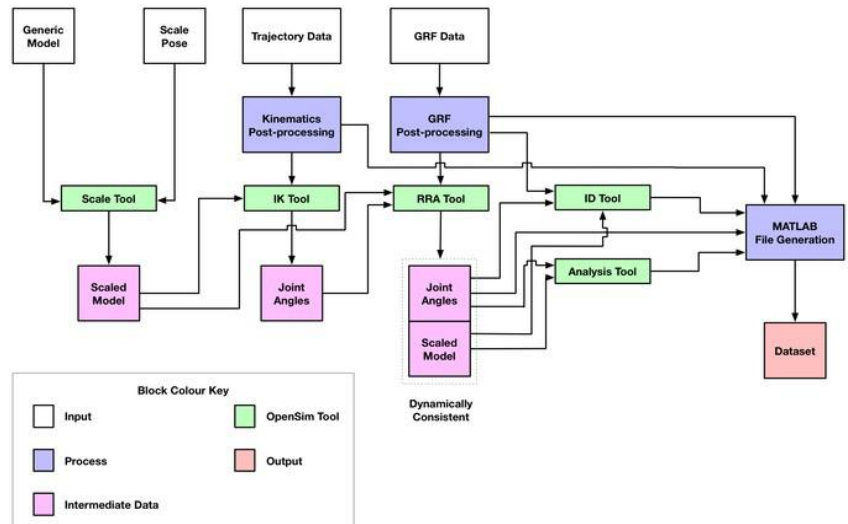

Figure 3.7: Schematic of data analysis pipeline.

Therefore, when computing acceleration, we decide to use another simple yet effective way in which we calculate the second order derivative of the position, based on the natural definition.

However, the position data are still a little noisy and taking the second order derivative of them will further increase the noise's impact, which can be seen in Figure 3.9a and 3.9b. It shows that the overall trajectory of the trial is evident while data points have clear fluctuation. Therefore, we decide to have some filtering to the data to clean up the noise and keep the trajectory's shape unchanged. We can infer that these fluctuations are high-frequency noises and the filter we apply cannot influence a trajectory's shape.

As a result we decide to use a fourth order Butterworth [7] low-pass filter with a cut-off frequency of 30 Hz. The reason is that Butterworth filter has a maximally flat passband (with no ripples) and its frequency response become zero (also without ripple) in its stop-band, compared with Chebyshev and elliptic filter types [52]. Thus signals in the passband will not be affected by the filter, and undesired noise in stop-band will all be filtered out. These parameters are set based on careful tuning while there can be other reasonable filter order and cut-off frequency combinations.

To trim the data, we go back to the VICON motion records to have a rough statistics and determine a time range relative to the treadmill's movement when the push recovery starts. Based on common knowledge, the push recovery should start at the point where the treadmill stops. While the maximum acceleration the treadmill has is $3m/s^2$ which cannot be traded as an instant stop, we inferred that the recovery should happen after the treadmill began to decelerate but before it completely stopped. However, based on the statistics of motion recordings, most the push recoveries started at the time when the treadmill started to decelerate, while some of them started even earlier than that time. Nevertheless, in both situations mentioned above, the data of the CoM's three properties are contaminated by the treadmill's behaviour, and in order to make the data only related to human motion, we decide to trim all trials in the way where the start of the trial is at the time when the treadmill's velocity reaches 0. By this way, we can be certain that the treadmill does not contaminate the CoM's position, velocity and acceleration in a trial. Moreover, for some trials, the participant has some unwanted motion after recovering to balanced which introduces irrelevant data. If this situation happens, we have another script to trim the data to deal with it manually.

Finally, after completing all calculations, we need to shift the CoM position data. The reason is that position data are recorded in the treadmill's coordinate which has the origin on the treadmill's centre. While to analyse the control behaviour, we need all three properties to be related to the CoM only. Therefore, we need to shift the data in order to make the origin be the CoM's position when the push start and thus have unbiased position data. An example of unbiased data can be seen in Figure 3.9c where the CoM is around the point $(0, 0, 0)$ when in steady state for the toe strategy. The same goes to the ankle strategy, but for stepping strategies, there will be some shift in the CoM's steady-state position while the steady-state velocity and acceleration are also 0. In addition, we add different tags to every trial, based on what recovery strategy is used and whether the trial starts with a clear countdown or randomly. Here we refer to the VICON motion recordings again, and some trials are marked 'abandoned' because we find the original motion unsatisfactory, i.e. the recording ends when push recovery is not finished, or the participant does not have a proper recovery motion.

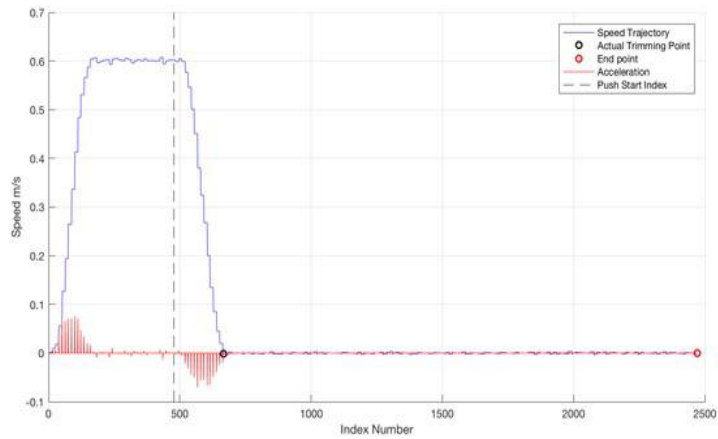

Figure 3.8: Trimming strategy demonstration on a diagram of treadmill speed

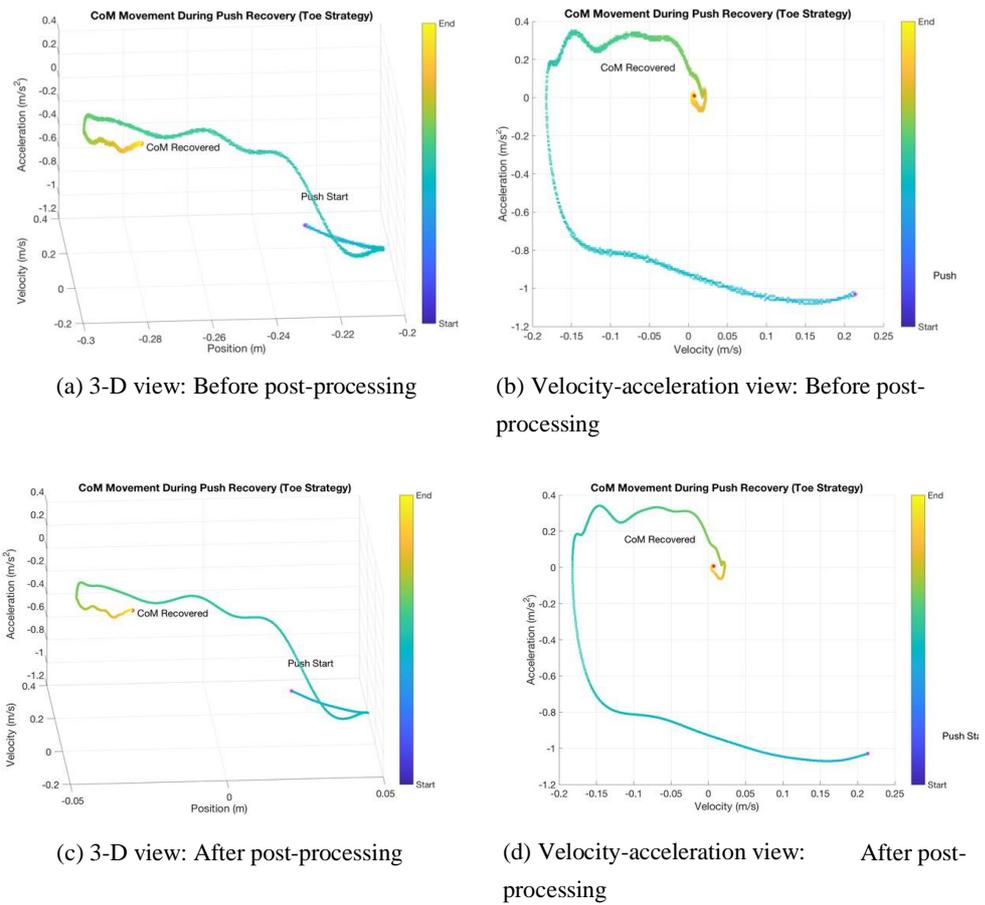

(a) 3-D view: Before post-processing

(b) Velocity-acceleration view: Before post-processing

(c) 3-D view: After post-processing

(d) Velocity-acceleration view: After post-processing

Figure 3.9: An example trial, before/after post-processing of filtering and position shift

## 3.3 Data Analysis

After all previous experiment, we gathered a dataset containing 323 valid trials from 7 participants. According to our observation, five kinds of strategies exist in this set of experiments, as presented in section 2.2. One example for each trial is shown in Figure 3.10 in 3-D view. Here we put the diagrams in velocity-acceleration view as well which is essentially one projection of the system's state space and this view gives a clear description of features which we use for reasoning and classifying different strategies. These 2-D diagrams are shown in Figure 3.11, and we can see clear features in the example of every strategy.

For ankle strategy, we can usually see a smooth rise of the trajectory followed by a turn which makes the acceleration of CoM change sign and finally return to the stable point, forming a hook shape. The CoM trajectory of the ankle strategy defines the fundamental trend of trajectories in all strategies in our scope of push recovery study. For toe strategy, we can see a slowly ascending trend followed by a sudden rise of the trajectory where the participant returns from standing on the tip-toe to standing on the sole. This feature differs toe strategy from others and is the determinant feature of this strategy. Moreover, trajectories of one-step strategy have a quick drop at the beginning which reaches a valley and starts ascending, and those of two-step strategy has another drop with another bottom which usually has smaller magnitude, following the first step. The toe-to-step strategy combines toe strategy and one-step strategy in series so we can see the rising feature in toe strategy followed by a complete one-step strategy trajectory.

We decide to use all 323 trials in further strategy selection statistics presented in sub-section 3.3.4. However, the model fitting part we present in subsection 3.3.1 and subsection 3.3.2, some trials are left out because we find that the corresponding trajectories in these trials are very messy which will only contaminate the fitted results. However, these messy trajectories' starting points are usually reasonable thus can be used in the statistics. Therefore, by leaving 50 trials out, we have 17 ankle strategy trials, 163 toe strategy trials, 14 toe-to-step strategy trials, 36 one-step strategy trials and 43 two-step strategy trials. The main reason for this number distribution is that the number of toe strategy is dominant during the experiment and we infer the reason to be our selected treadmill speed range.

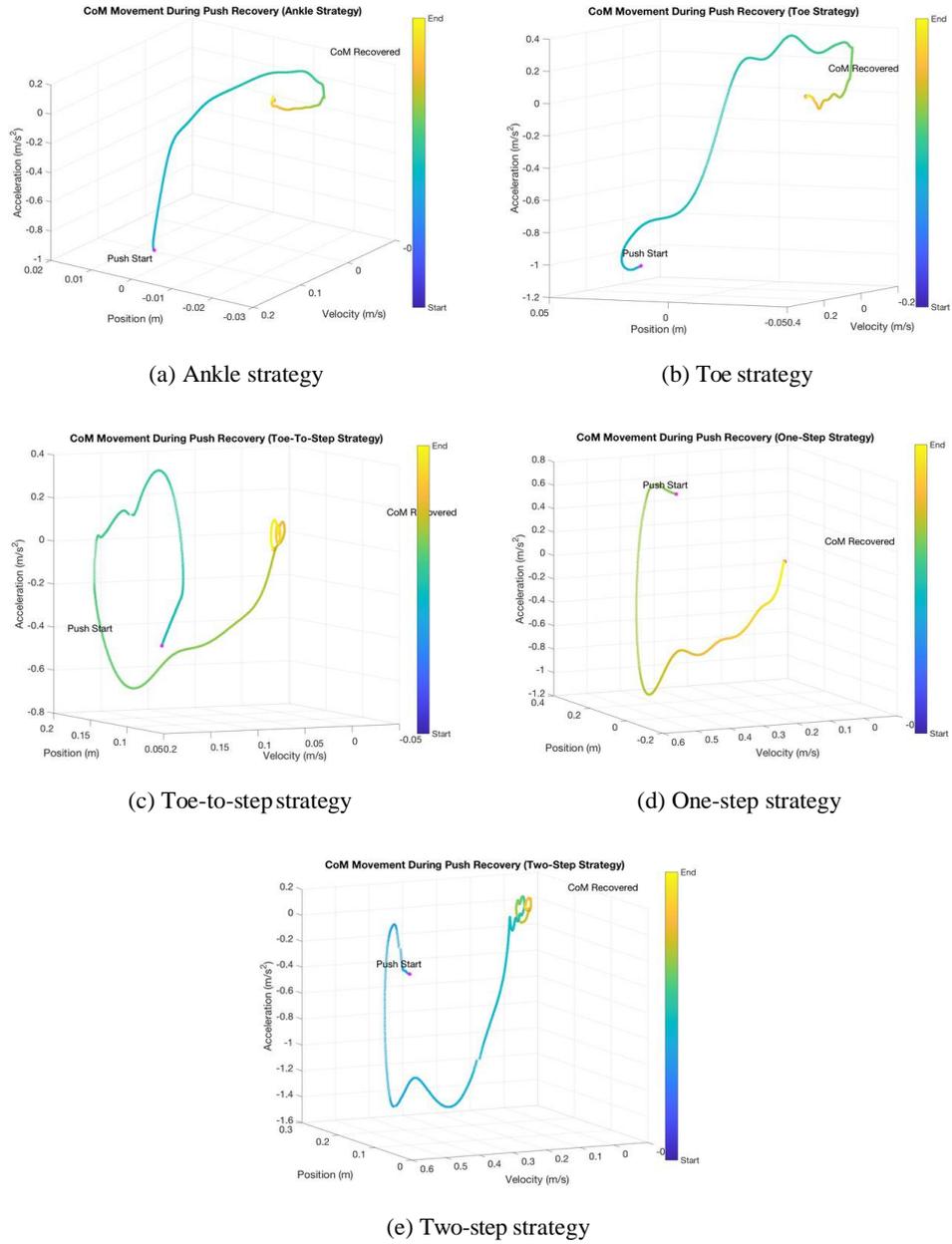

Figure 3.10: Example trials of every recovery strategy in 3-D view

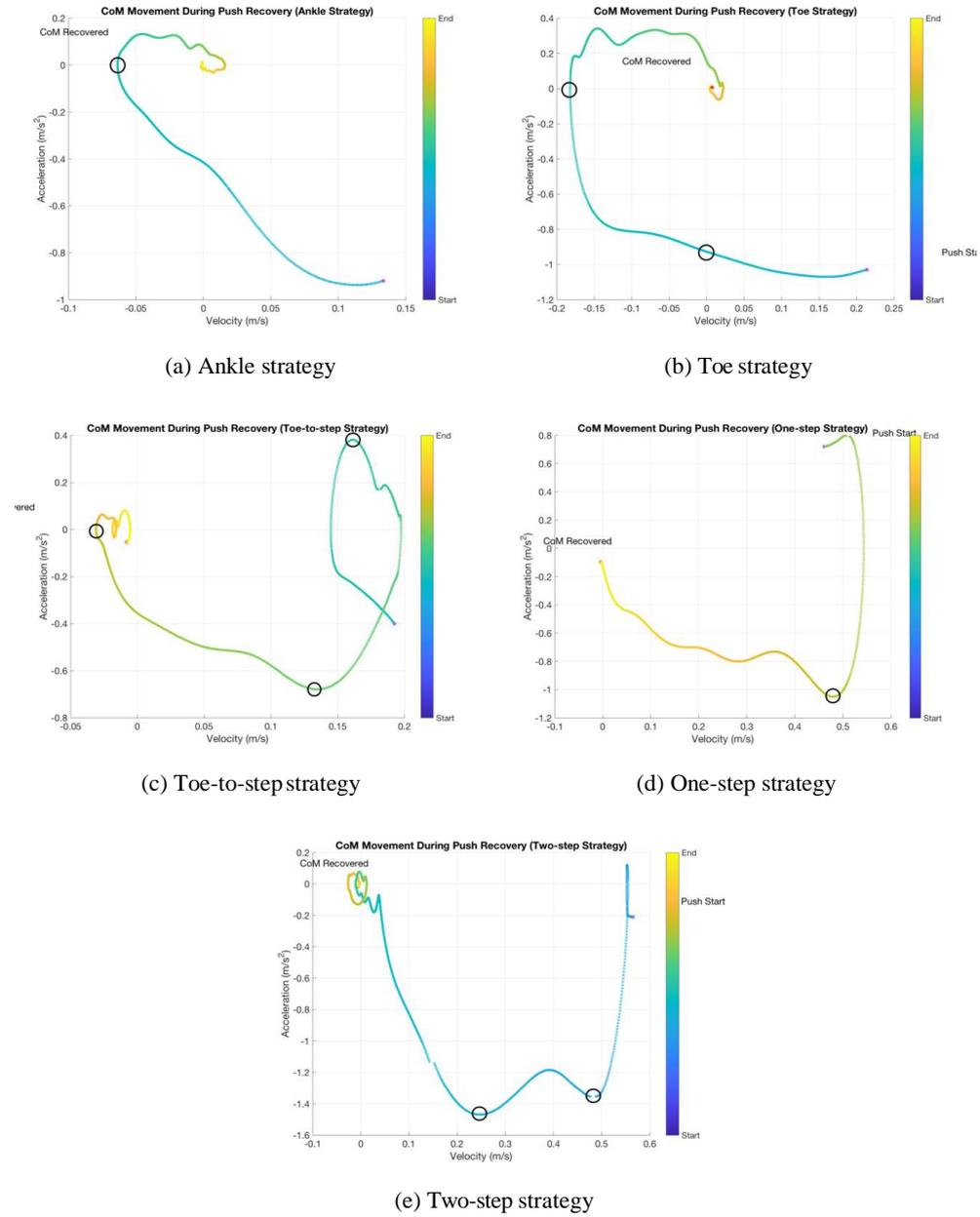

Figure 3.11: Example trials of every recovery strategy in velocity-acceleration view

### 3.3.1 Model Fitting Methodology

The model fitting is based on linear regression method. Because control gains are usually tuned in practice, mixing all trials and having a fixed set of gains fitting may not be the most appropriate way. Therefore, we decide to fit each trial separately and do statistics on the outcomes to explore the controller gain variation properties.

We use CoM positions, velocities and accelerations to fit models based on PID control law as well as some of its subsets. The models we use for fitting in this part are P, PD, PI and PID control models, based on previous 1-D point mass push recovery definition and inferences in subsection 2.1.2 as well as the fact other controllers expect these four are usually not used. Although applying P control law alone to our model does not make sense when combining some inductions from subsection 2.1.1, we still include it in our scope to see its real effects when used solely. The formulation will be the discrete version of that of a 1-D point mass, and the modelling procedure also follows the contents in subsection 2.1.1. E.g. for PID control, the model is:

$$a_k = \frac{K_p}{m}(p^* - p_k) + \frac{K_i}{m}\sum_{j=1}^{k}(p^* - p_j) + \frac{K_d}{m}(v^* - v_k) \tag{3.1}$$

where $a$ is the CoM's acceleration and $p$, $v$, $m$ are the CoM's position, velocity and mass. The reference point $(p*, v*)$ is the CoM's position and velocity in steady state. Although by definition the velocity should be 0, here we take the last data point of a trial as the reference point when the subject is fully recovered while the calculated CoM state may not be the ideal steady state. Nevertheless, the last point of most trials have small enough values such that they only influence the results a little. All other PID control law subsets follow the same formulation pattern. No bias term is included in any of the models we are going to fit with because of the steady state definition by which the acceleration will become 0 if all error signals reach 0. On the other hand, we reckon that a bias term lacks some physical meaning to be included in the formulation.

For the preliminary analysis, we use all data points in one push recovery trial to fit one model. After that, locally weighted regression [9] for different segmentations of a trial, as well as control error metric exploration are done. These two parts are presented in subsection 3.3.2 and subsection 3.3.3 respectively.

To be noticed that, in this project, we essentially have only the training set since every trial is fitted separately. However, it is possible that for some trials, fitted parameters have extreme values that will influence further statistics with outliers introduced. Trying to prevent the outliers' effect, we decide to use the regularisation technique [44] in the regression process to punish extreme coefficients. In this way we sacrifice some accuracy to get a more reasonable set of parameters for the model, hence reduce the outliers' effect. By adding a regularisation term, the least square cost function being minimised becomes:

$$E_\lambda = \sum_{i=1}^{N}(y_i - f_i)^2 + \lambda \sum_{k=1}^{K} w_k^2 \tag{3.2}$$

Here, $f_i$ represents a predicted value calculated by the model, while $y_i$ is the corresponding true observed value. $w_k$ represents a fitted parameter and $\lambda$ is the regularisation constant. In order to accomplish this cost function in normal linear regression, we change the input and output vectors by adding new data items:

$$\mathbf{y}' = \begin{bmatrix} \mathbf{y} \\ \mathbf{0}_k \end{bmatrix}, \qquad \mathbf{x}' = \begin{bmatrix} \mathbf{x} \\ \sqrt{\lambda}\mathbf{I}_k \end{bmatrix} \tag{3.3}$$

The two new terms assume the size of the input feature matrix $\mathbf{x}$ is $N \times K$, and the size of the observed-value vector $\mathbf{y}$ is $1 \times K$. Here $\mathbf{0}_k$ is a vector of $K$ zeros and $\mathbf{I}_k$ is a $K \times K$ identity matrix. By doing the above input-output modification, we apply regularisation to normal least square linear regression. Within the scope of this project, we choose a regularisation constant of $\lambda = 0.01$ which can penalise extreme parameters a little while keeping the accuracy as unaffected as possible.

To evaluate the fitted models, we use the coefficient of determination ($R^2$ metric) [16] as well as the root mean square error. Some reasoning in the model's parameters is also done during the result evaluations. The $R^2$ metric is defined as Equation 3.4, which normally gives the percentage of the dependent variable variability which has been taken into account. If the calculated $R_2$ metric is negative, it no longer represents the percentage of the dependent variable variability and means the model needs a bias term. Since we cannot have a bias term by definition, we can judge that a negative $R_2$ metric suggests a lousy model choice.

$$R \equiv 1 - \frac{S_{res}}{S_{tot}} \tag{3.4}$$

Here in Equation 3.4, $S_{res}$ is the residual square error, and $S_{tot}$ is the total sum of square, defined by the following equations:

$$S_{\text{res}} = \sum_i (f_i - y_i)^2 \tag{3.5}$$

$$S_{\text{tot}} = \sum_i (y_i - \bar{y})^2 \tag{3.6}$$

$$\bar{y} = \frac{1}{n}\sum_i^n y_i \tag{3.7}$$

In these equations, $f_i$ represents a predicted value calculated by the model while $y_i$ is the corresponding observed value with the mean being $\bar{y}$. Apart from all equations to get the $R^2$ metric, the root mean square (RMS) error calculation is done using Equation 3.8 with all notations being the same as the above equations. The RMS error is used to judge the performance of the fitted model given a set of input and output.

$$RMS_{error} = \sqrt{\frac{\sum_{i=1}^n (f_i - y_i)^2}{n}} \tag{3.8}$$

All model-fitting related analysis of the project is based on the two metrics and output parameters presented in this subsection. In addition, we use the MATLAB function 'mldivid' [38] for our model fitting which naturally has no constraints on coefficients. However, since controller gains are defined to be positive numbers, having a negative result shows that the fitting is not reasonable.

### 3.3.2 Model Fitting on Trial Segmentations

In the first set of model fitting, we use full trials in the dataset. Doing this suggests that during a push recovery trial a person uses a fix set of control gains, while this presumption does not necessarily need to be true. Possibility lies in people having different gains for different phases of a push recovery trial when using different recovery strategy. In this subsection, we present the second set of model fitting, to fit on trial segmentations. Based on our observation, every recovery strategy has significant motion milestones, which are consistent among all participants. Therefore, we can segment trials of each recovery strategy using these natural hierarchical motions.

For ankle strategy, we can segment one trial into two parts at the point where acceleration changes its sign for the first time, i.e. from negative to positive. The first part is hence from the start of the trial to the first time when the CoM goes across steady-state point (subject leaning backwards and across the regular pose). The second part is leaning forward again to recover balance. There can be multiple crossings of the steady-state point, but we assume they share the same set of gains since the motions are much more moderate.

For toe strategy, we can segment one trial into three parts, and the first part is from trial start to the point where velocity first reaches 0, indicating that the person is standing on the top of the tip-toe. The second part is from this point to the time when the person stands on full-sole again with the acceleration reaches 0, and starts to lean backwards. The last part is the same as that of the ankle strategy, which is leaning forward to recover to steady-state pose. Finally, for one-step strategy, we can also split one trial into two parts. The first part is making a step and the second one is full body leaning until recovering to normal pose after the step. Since the two combined strategies share the same set of motion milestones thus we decide not to list them separately. All milestones have been marked with black circles shown in Figure 3.11 for a clear observation.

Based on our reasoning, a push recovery trial can be treated as combinations of hierarchical motion primitives, with different strategies indicating different combinations. According to the above segmentation decision, we can define three kinds of hierarchical motion primitives: leaning forward/backwards, lifting/dropping the feet to stand on tip-toe/sole and making a step. Hence, we can also infer that a trial of the two-step strategy combines two stepping phases and a full body leaning phase. The toe-to-step strategy is a combination of lifting to tip-toe and making a step, where sometimes the switch of the action happens before the previous movement finishes.

Based on these inductions, we choose to fit on segmentation of the ankle, toe, and one-step strategy trials in this section because they contain one set of motion primitives respectively. For combined strategies, since they have one type of preliminary movement multiple times, we should be able to fit them by tuning the gains of corresponding preliminary movements, according to which we decide to leave them out. In this stage, we use PD and PID control laws to explore how the overall outcomes vary when fitting on segmentations, and whether the behaviour of the integral term influences the performance. When using integral control, we still keep track of accumulated errors for all previous time steps. Although we need to fit models on segmentations here, we believe that this accumulated error should be inherited in further segmentation of a trial because it makes more sense to keep it consistent rather than resetting it for every phase based on the fact that people should remember what has been done.

### 3.3.3 Error Metric Exploration

Until now we use classic PID control law family to fit our model. These control laws all use linear error being the difference between the target value and current value, e.g. $e = p^* - p$. Here we would like to explore whether changing the error metric can have any improvement in the model's performance in model fitting. We decide to use an odd order polynomial because using even order will make the error lose its sign while in the field of control the sign of error is also needed to get to the steady state. However, as an exploration, we also change the original linear error to exponential scale and see what impact it will bring to the model, which can also help us explore the effect of the signal's sign since exponential values are all positive. In addition, according to the results from previous experiments, we decide to use the original PD control model as the baseline. Therefore, in this part of the analysis, we have two error metrics to explore, and the control formulations are:

$$a = \frac{K_p}{m}(e^7 + e^5 + e^3 + e) + \frac{K_d}{m}((e^7)' + (e^5)' + (e^3)' + \dot{e}) \tag{3.9}$$

and

$$a = \frac{K_p}{m}\exp(e) + \frac{K_d}{m}\exp(\dot{e}) \tag{3.10}$$

Here, $e$ represents the original linear error metric and the $(e^x)'$ form represents the differentiation of that error signal, with $\dot{e}$ being the first order derivative. To be noted that in this part of the experiment, all other settings are kept the same as previous analysis phases.

### 3.3.4 Strategy Selection Statistics

In addition to control law, we are also interested in how people select push recovery strategies. In previous studies, B. Stephens *et al*. has studied push recovery strategy selection in humanoid robots with feedback controls on LIPM models [45]. The simulation results give stability regions of the CoP's initial position and velocity. This inspires us to have a statistics of CoM's initial position and velocity of every valid trial and observe how is strategy selection related with initial states. Although due to the trimming, the initial states of the trials are after the push because of the real starting profiles being contaminated by the treadmill, we can still infer qualitatively. The reason is that the velocity should be a little faster while the position should be closer to the origin at the real start of the push. Therefore, most dots in the statistics plot for real push start should be shifted in uplift direction while the inter-relationships distributed point are still roughly the same. All analysis details of this part are presented in section 4.4.

# Chapter 4

# Results and Discussion

This chapter presents results and our discussion based on them. There are four sections in this chapter corresponding to the four data analysis phases presented in section 3.3, and we present them in the same order here.

## 4.1 Full-trial Fitting

In this section, we present results of our primary analysis, model fitting on full trials. We split the section into three subsections, and in the first two of which we mainly present graphical examples of fitted models and original trajectories to give qualitative inductions. In subsection 4.1.3, we give a statistics result on fitted parameters and evaluation metrics for all trials which help give quantitative results.

### 4.1.1 Proportional(P) and Proportional-Derivative(PD) Control Laws

Starting from the simplest control law in our scope, we have model fitting with only P control law. This essentially reduces the state space to 1-D and the overall relationship into 2-D. According to the plots of position-acceleration for P control law, we can see the corresponding fitted models are actually a straight line shown in Figure 4.1 which cannot capture any detailed features. This pattern is consistent in all trials and fits the original behaviour of the 1-D point mass dynamics since no velocity constraint is added to the system, the overall control leads to 1-D position oscillation around the steady-state point. Therefore, we can summarise that P control law cannot describe the relationship between the control applied and the CoM's current state based on the graphical results.

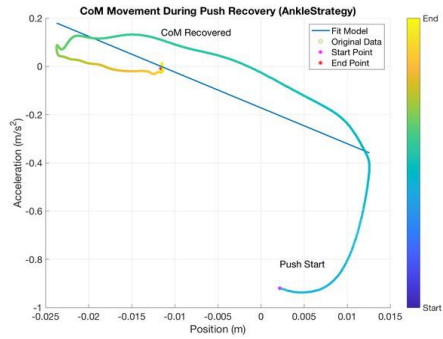

(a) Ankle strategy

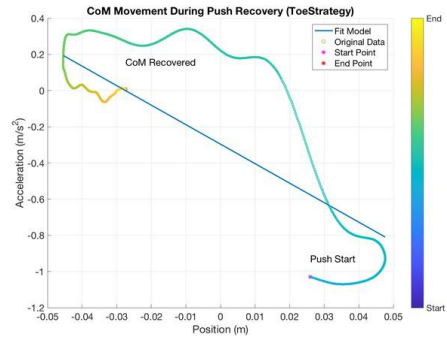

(b) Toe strategy

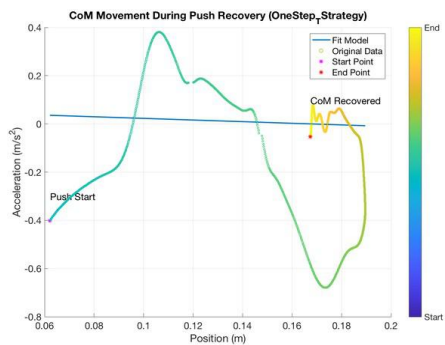

(c) Toe-to-step strategy

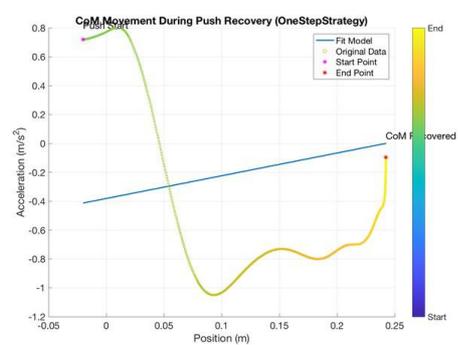

(d) One-step strategy

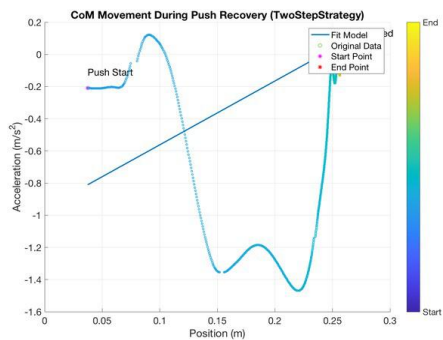

(e) Two-step strategy

Figure 4.1: Example trials of all strategies, fit with P control law on full-trial

Having seen the effects only using P control law, we now fit all these strategies with PD control which gives a 2-D state space and 3-D overall relationship as shown in Figure 3.10. The result diagrams are shown in Figure 4.2. The black transparent surfaces represent the working domain of the fitted PD controller, and the 'Fit Model' lines are the directly fitted model of the trial. Also, we include both position-acceleration and velocity-acceleration view for a clearer 2-D observation with respect to each input although the fitted model is related to both of them, and plots in position-acceleration and velocity-acceleration views can be found in Figure C.1.

It is clear that according to Figure 4.2, models fitted with PD control law perform much better than those fitted with P control. PD control fits trials of ankle strategy well by capturing almost all trends while there are some small gaps between the fitted and the original trajectories, and there seems to be an offset between the working domains and original trajectories. For toe strategy, PD controller also captures trajectories of trials with its working domain since the planes describe the trajectories well. However, some 3-D trends in toe strategy trials are ignored because of the linearity of state space relationship in PD control. However, for strategies having more complicated trends, it starts not to take more details into account but to only try to capture the major trends. The clearest example can be the one of toe-to-step strategy shown in Figure 4.2c which is various in the 3-D space. One 2-D plane is not capable of capturing all these features. Nevertheless, combining P and D terms gives us better model fitting results by being able to capture the most significant trajectory trend. This finding also reinforces our hypothesis to some extent.

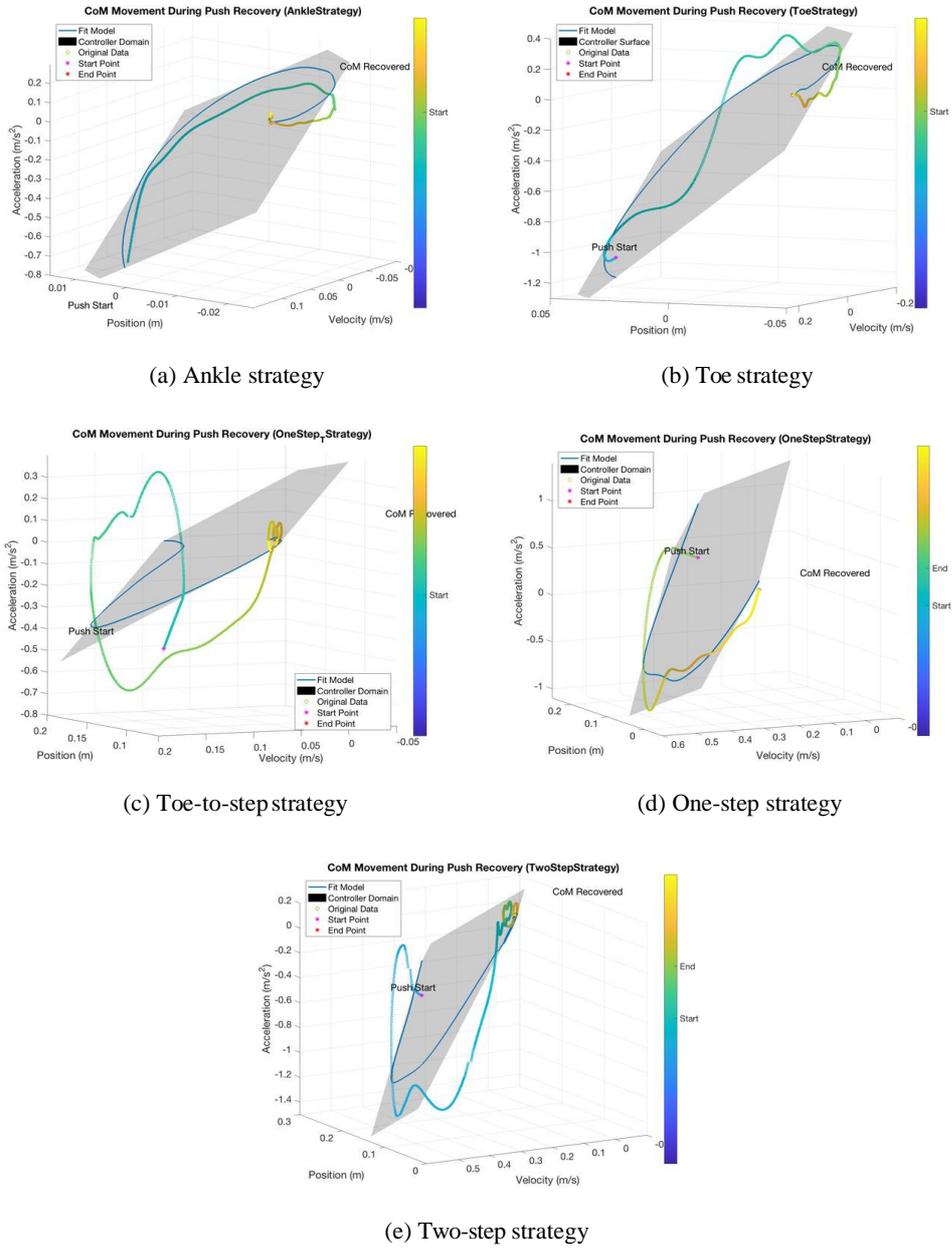

Figure 4.2: Example trials of all strategies in 3-D view, fit with PD control law on full-trial

### 4.1.2 Including the Integral Term

So far our results give us an intermediate conclusion that PD control is able to describe the major trend of a push recovery trial. In this subsection, we try to explore how adding the integral (I) term to our model fitting influences the overall results. To begin with, we try PI control, and the graphical results for example trials can be seen in Figure 4.3. We find a general pattern here that having the integral term helps the P control by adding some non-linearity to the fitted model. However, PI control is still not able to describe the relationship between the input feature and output control (acceleration), and we can now confirm that both P and D terms are necessary to properly describe the control relationship using the 1-D point mass model of the CoM.

We then apply the PID control law and figures of the same set of example trials are shown in Figure 4.4, and plots in position-acceleration and velocity-acceleration views can be found in Figure C.2. We can see the controller domains now become 3-D surfaces because of the non-linearity introduced by the I control, while the major parts of these domains are still 2-D planes. These diagrams also show that fitted model of the ankle, toe-to-step and two-step strategies become closer to the original trajectory using PID control while that of toe-to-step strategy trial does not change sufficiently to better describe the trial. However, we cannot see any apparent changes to those of toe and one-step strategies although we see the controller domains have clear 3-D features in toe strategy trials. On the other hand, all models fitted with PID control are still not able to capture all features presented in subsection 4.1.1 with the toe-to-step strategy trial being the most explicit example again. However, to see whether having the integral term makes the model better quantitatively we need to refer to the values of parameters, which we are going to present in the next subsection.

### 4.1.3 Result Statistics

Here we give a result statistics table of all strategies fitted with all models for every participant. We decide to give the average value for fitted parameters as well as evaluation metrics of all trials having the same recovery strategy. Moreover, the **mean absolute deviation** (MAD) is also presented in the statistics to get better insights into variations of parameter values. Results for the full-trial fittings are shown in Table 4.1. In the table we have both parameter values fitted with PD and PID control laws and fitted controller gains are given in normalised gains (normalised by the participants' mass).

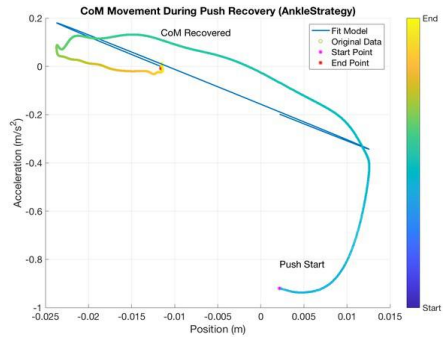
(a) Ankle strategy

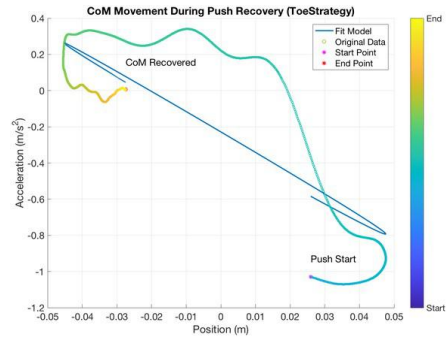
(b) Toe strategy

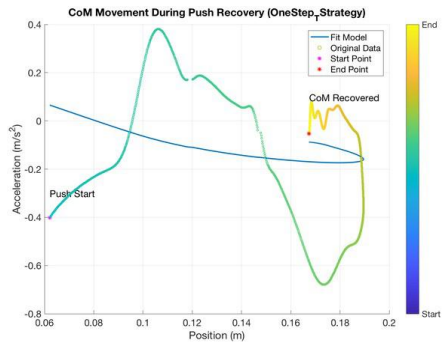
(c) Toe-to-step strategy

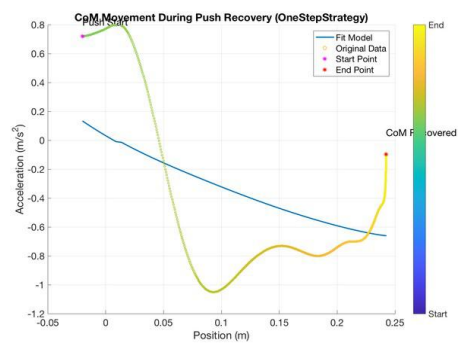
(d) One-step strategy

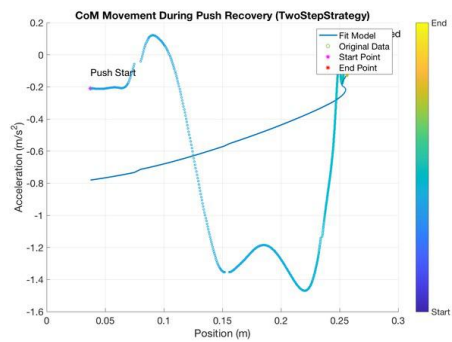
(e) Two-step strategy

Figure 4.3: Example trials of all strategies, fit with PI control law on full-trial

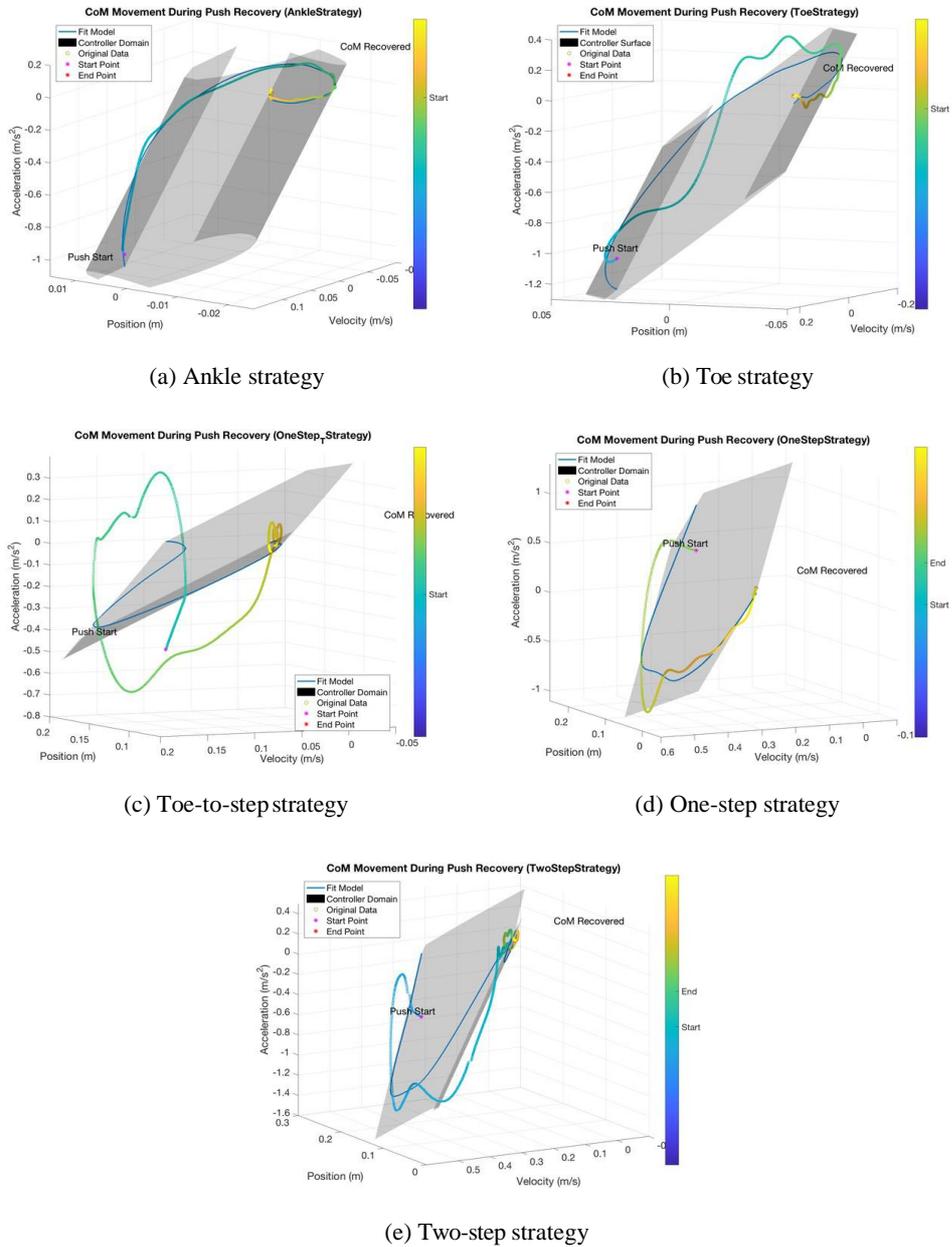

Figure 4.4: Example trials of all strategies in 3-D view, fit with PID control law on full-trial

According to the RMS errors, both PD and PID outcomes of ankle and toe strategies are within 0.1 while those of the other three strategies go above 0.1 but below 0.2. For $R^2$ metrics, all results except those of toe-to-step strategy reach above 0.8, indicating more than 80% of the input variables are considered during model fitting. Comparing these two kinds of evaluation metrics, we find that PID control manages to have better performances by reducing the RMS error by at most 0.025 of corresponding PD results. $R^2$ metrics of PID outcomes are also at most 0.02 higher than those of PD control. However, the two evaluation metrics are entirely the same when it goes to toe and one-step strategies, where having the integral term does not improve the model by any means.

Therefore, we turn to the three fitted gains for further induction. We can see that average values of proportional (P) and derivative (D) gains have some differences in the two control laws while the differences are all smaller than 1. However, using PID control increases the MAD of these two parameters excluding those of the one-step strategy. When referring to the integral gain, we can see all integral gains are considerably (more than 100 times) smaller than the other two gains and this pattern is consistent in all recover strategies. Hence the integral term does not have a real effect in its corresponding control formulation but giving the formulation higher fault-tolerance. Based on these findings, we summarise that having the integral term makes the model more complicated to take more delicate trajectory pattern and thus more data points into account. This also leads to better model fitting outcomes by having smaller RMS errors. Moreover, since the average $K_i$ values of toe-to-step and one-step strategies are negative, we infer that using the PID formulation does not lead to reasonably fitted model at least in these two strategies since all gains should be positive. Therefore, we conclude that within our project scope, PD control is the control law which can describe the relationship amongst our collected CoM properties for full-trial model fitting, with average RMS errors around 0.127 and 82.3% independent variables considered on average, giving each strategy the same weight. However, since all other strategies have much fewer trials compared with the toe strategy according to section 3.3, these two average values may not indeed represent our models' performance.

Table 4.1: Result statistics of fitted controller gains and evaluation metrics, full-trial fitting

(The upper number in a cell is for PD control while the lower one is for PID control)

| Strategies \ Parameters | $K_p$/m | | $K_i$/m | | $K_d$/m | | RMS Error | | $R^2$ Metric | |
|---|---|---|---|---|---|---|---|---|---|---|
|  | Mean | MAD | Mean | MAD | Mean | MAD | Mean | MAD | Mean | MAD |
| Ankle | 8.0239 | 3.0673 | —— | —— | 3.5963 | 0.6527 | 0.0819 | 0.0479 | 0.8722 | 0.0672 |
|  | 7.9180 | 3.5742 | 0.0021 | 0.0048 | 3.5911 | 1.1514 | 0.0556 | 0.0217 | 0.8977 | 0.0612 |
| Toe | 8.3515 | 2.8139 | —— | —— | 2.5100 | 0.7536 | 0.0849 | 0.0309 | 0.8681 | 0.0749 |
|  | 8.7603 | 2.9524 | 0.0010 | 0.0034 | 2.7157 | 0.8387 | 0.0849 | 0.0309 | 0.8681 | 0.0749 |
| Toe-to-step | 5.3920 | 4.1288 | —— | —— | 2.7340 | 1.2988 | 0.1639 | 0.0372 | 0.7247 | 0.0788 |
|  | 5.7658 | 4.4749 | -0.0014 | 0.0010 | 2.2054 | 1.3729 | 0.1468 | 0.0366 | 0.7459 | 0.0644 |
| One-step | 12.7519 | 5.5033 | —— | —— | 5.4339 | 2.2588 | 0.1661 | 0.0714 | 0.8251 | 0.0947 |
|  | 11.8275 | 5.1234 | -0.0042 | 0.0078 | 4.8585 | 1.7459 | 0.1661 | 0.0714 | 0.8251 | 0.0947 |
| Two-step | 10.8154 | 5.3315 | —— | —— | 4.8350 | 0.6658 | 0.1384 | 0.0431 | 0.8286 | 0.0785 |
|  | 11.1408 | 5.3837 | 0.0001 | 0.0046 | 4.8579 | 0.7019 | 0.1290 | 0.0448 | 0.8487 | 0.0799 |

## 4.2 Segmentation Model Fitting

In this section, we present the results of model fitting on segmentations. Based on previous results we decide to fit models with only PD and PID control laws in this phase. Based on our reasoning back in subsection 3.3.2, the results presented here are trials containing original motion primitive sets. To be specific, we use trials of ankle, toe and one-step strategies to get the results. The two combined strategies are not included in this part of the study. The fitted models and original trajectories are shown in Figure 4.5 and Figure 4.6 for PD and PID fittings respectively. For observation ease, we put two separate diagrams in different views for the toe-strategy example trial. Also, we include the fitted models' examples in the position-acceleration and velocity-acceleration views in appendix figures, Figure C.3 and Figure C.4, the same as what we do in the previous section.

According to these diagrams, both PD and PID control laws fit backwards-to-forward leaning movements well in all three strategies (Fit Model 2, 3, and 2 of the ankle, toe, and one-step strategy trials). Also, we do not have visually significant differences in the fitted models of ankle and toe strategies despite the non-linearity introduced by I control in this motion primitive. Similar goes to the fittings on forward-leaning of ankle strategy and toe-lifting phase of toe strategy, while models fitted with PD control seems to be more natural in trends compared with those fitted with PID control, and 3-D surfaces created by the PID controller seem to be redundant describing these two motions. Moreover, based on the controller domains and how they fit the trajectory, we can conclude that segmenting a trial based on hierarchical short-term motion primitives can be a reasonable reference to change controller gains since we have clearly better fits that planes can describe the original trajectories in a more natural way.

In addition, PID control seems to have better-fitted outcomes in movements of making a step since the created 3-D surface is able to contain the majority of data points for this motion and the fitted model's shape follows the trajectory well. As for dropping from tip-toe to heel, although we get better fittings with PID control by having the shape of trajectories better described, neither PID and PD control can capture the major feature trend of this action.

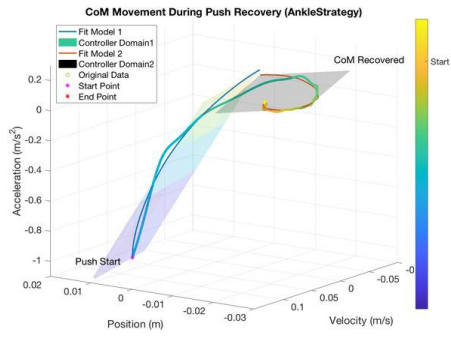
(a) Ankle strategy

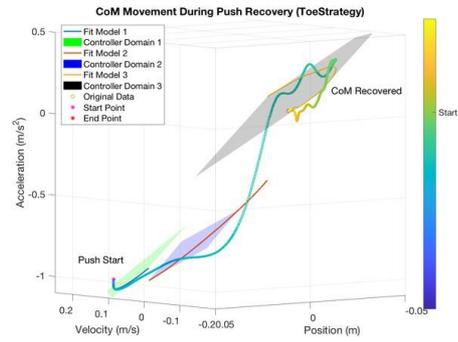
(b) Toe strategy, one view

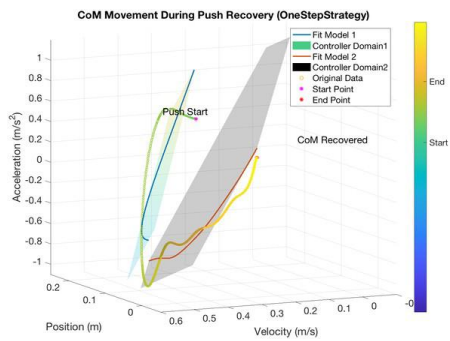
(c) One-step strategy

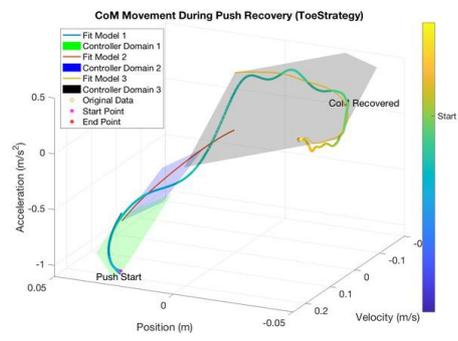
(d) Toe strategy, another view

Figure 4.5: Example trials of the three selected strategies in 3-D view, fit with PD control law on segmentations

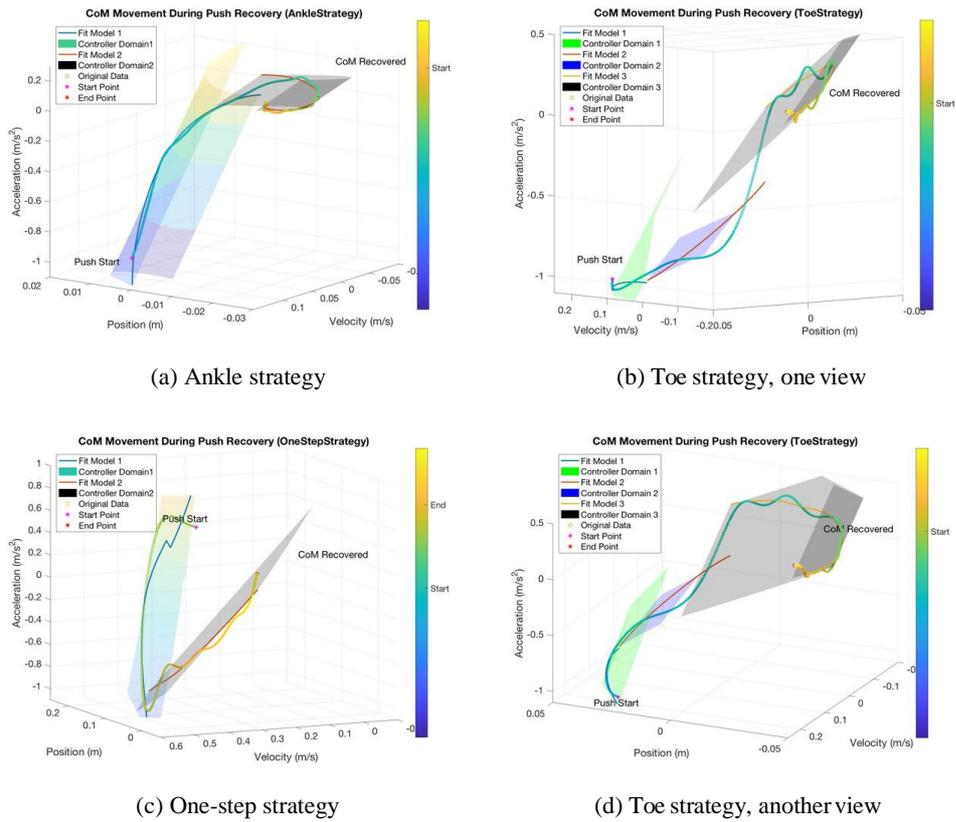

(a) Ankle strategy

(b) Toe strategy, one view

(c) One-step strategy

(d) Toe strategy, another view

Figure 4.6: Example trials of the three selected strategies in 3-D view, fit with PID control law on segmentations

### 4.2.1 Result Statistics

Now we turn to the actual numbers for quantitative analysis. As it is in subsection 4.1.3, average values and mean absolute deviations of every parameter can be seen in Table 4.2. Because we have a classification of movement primitives, we decide to do the statistics based on these movements. Therefore, the table is built up with the three chosen strategies in different phases. For ankle strategy, phase 1 is leaning forward and backwards during which the CoM goes across the steady-state point, and phase 2 is leaning backwards after the crossing of the steady-state point and then forward again until covered to balanced. For toe strategy, phase 1 is to lift heels and stand the tip-toe, while phase 2 is dropping the heel to stand on full sole again until the CoM goes across the steady-state point. Moreover, phase 3 is leaning forward again until recover. Finally, phase 1 for one-step strategy is making a step, and phase 2 represents full body leaning after finishing making the step. Besides, RMS errors and $R^2$ metrics here are given weighted average values for the whole trial based on data point proportions.

According to the table, all $R^2$ metrics are much higher than those in previous full-trial model fitting section in Table 4.1, indicating more data points are taken into account when fitting the model. On the other hand, RMS errors for both ankle and toe strategies are smaller when fitting on segmentation, the decrement is more than 25% compared with full-trial fitting results. However, for one-step strategy, this value increases for fitting on segmentations. The reason can be that fitted models of this strategy's phase 2 have larger RMS errors because phase 2 has much more data points. During full-trial fitting, models are set with larger weight to fit on points in this phase 2 portion while sacrificing points in phase 1 to have lower RMS errors.

When referring to fitted gains, we can see significant differences in proportional gains between different phases given a recovery strategy, e.g. phase 1 and 2 of ankle strategy have proportional gains with magnitude differences larger than 5. The same goes to phase 1, 2 and 3 of toe strategy, as well as phase 1 and 2 of one-step strategy. This finding reinforces our hypothesis that people may have different sets of gains during a push recovery trial. Moreover, we find that the gain differences of proportional gains are more considerable than those of derivative gains. On the other hand, integral gains in this part of study have a similar pattern as the one in subsection 4.1.3, yet for some actions, e.g. phase 1 of one-step strategy, using PID control also makes average values of $K_p$ negative, suggesting an unreasonable model choice. Hence, we may draw the same conclusion that adding the integral term does not really help improve the model's ability to fit push recovery processes. Therefore, we can summarise that PD control can describe push recovery with different gains, set for different phases of a trial. The average RMS error is 0.09, and 93.7% of data points are taken into account on average which are both better than those of full-trial fitting.

However, none of the values presented above can help us reason why fitted models of toe strategy phase 2 and one-step strategy phase 1 are not able to capture the primary trend of corresponding trajectories. Our explanation is that gravity is involved in both of these movements which introduces extra kinetic energy throughout the motions. It is hard to tell how much of the behaviour is caused by the gravity compared with human control while our model is fitted based on the assumption that all motions are done in full human control. Besides, during these two movements the CoM has more significant movements in the vertical direction (perpendicular to the model's dimension), so extending the model into 3-D may help reveal the truth.

Table 4.2: Result statistics of fitted controller gains and evaluation metrics, segmentation fitting

(The upper number in a cell is for PD control while the lower one is for PID control)

| Strategies \ Parameters | $K_p$/m | | $K_i$/m | | $K_d$/m | | RMS Error | | $R^2$ Metric | |
|---|---|---|---|---|---|---|---|---|---|---|
| | Mean | MAD | Mean | MAD | Mean | MAD | Mean | MAD | Mean | MAD |
| Ankle, phase 1 | 8.8606 | 8.1459 | —— | —— | 3.2908 | 1.5161 | | | | |
| | 6.8308 | 6.3230 | 0.0432 | 0.1322 | 4.3445 | 2.4409 | 0.0495 | 0.0281 | 0.9341 | 0.0654 |
| Ankle, phase 2 | 3.4625 | 2.0586 | —— | —— | 1.5804 | 0.7897 | 0.0419 | 0.0238 | 0.9477 | 0.0569 |
| | 3.6424 | 2.0169 | -0.0009 | 0.0010 | 1.1150 | 0.8352 | | | | |
| Toe, phase 1 | 9.2857 | 3.8510 | —— | —— | 3.3103 | 2.1023 | | | | |
| | 2.1966 | 1.5157 | 0.0006 | 0.1403 | 4.3201 | 1.3317 | | | | |
| Toe, phase 2 | 9.9908 | 3.8436 | —— | —— | 1.9176 | 1.6895 | 0.0473 | 0.0180 | 0.9888 | 0.0149 |
| | 8.1633 | 3.2894 | 0.0456 | 0.0707 | 3.4519 | 2.5490 | 0.0418 | 0.0160 | 0.9952 | 0.0064 |
| Toe, phase 3 | 4.2651 | 2.8199 | —— | —— | 2.1680 | 0.7639 | | | | |
| | 4.9930 | 3.002 | -0.0003 | 0.0011 | 2.2503 | 0.9612 | | | | |
| One-step, phase 1 | 13.5108 | 6.8657 | —— | —— | 5.2996 | 3.1030 | | | | |
| | -1.5583 | 3.8934 | -0.0810 | 0.1019 | 0.1639 | 3.6128 | 0.1761 | 0.0600 | 0.8900 | 0.0712 |
| One-step, phase 2 | 7.5537 | 4.9176 | —— | —— | 4.4279 | 1.6842 | 0.1235 | 0.0375 | 0.9738 | 0.0341 |
| | 12.1181 | 6.7379 | -0.0042 | 0.0066 | 3.5142 | 1.3150 | | | | |

## 4.3 Error Metric Exploration

During the implementation of this stage's experiment, we fit models for some particular trials of each recovery strategy with the two different error metrics. If the models have any better performances, it is then generalised to all trials. To present the discovery, we have two sets of results from example trials of ankle and toe strategies. The results are presented in two tables, Table 4.3 and Table 4.4, for example ankle and toe trials. For polynomial parameters in both tables, the 4 parameters in $K_p$ cells are gains of $e^7$, $e^5$, $e^3$ and $e$ terms. The same pattern goes for those in $K_d$ cells.

According to the two tables, we can see that exponential error metrics make the overall model performances much worse by having more than doubled RMS errors and very low $R^2$ metrics while that in the toe trial is negative, suggesting the model does not actually work correctly. Hence having exponential error metrics decreases the control model's performances, reinforcing our induction in subsection 3.3.3, and proving the idea that signs of signals does matter in control.

For the proposed polynomial error metric, it is clear that higher order terms (power of 5 and 7) have negligible gains while linear terms are almost identical to those of the first error metric, our baseline. Moreover, because the RMS errors and $R^2$ metrics of these two models with polynomial errors also have ignorable differences compared with the original ones, we can summarise that having polynomial error metrics cannot help improve model performances either. The tiny improvements are due to the fact that more complicated model has higher data-describing ability to get better fits. This situation is similar to the integral term's effect presented in previous sections. We have also tried other trials of different strategies, but the results are in the same pattern as of the examples given here thus no further unnecessary details are given here. Therefore, this subsection concludes that both of our proposed error metrics failed to improve the performance of our model.

Table 4.3: Results of example ankle strategy

(From top to bottom, values in the $K_p/m$ and $K_d/m$ cells of 'Polynomial' are gains for $e^7$, $e^5$, $e^3$ and $e$ terms)

| Strategies \ Parameters | $K_p/m$ | $K_d/m$ | RMS Error | $R^2$ Metric |
|---|---|---|---|---|
| Linear (Baseline) | 15.4599 | 3.8046 | 0.0854 | 0.8339 |
| Exponential | -1.4786 | 1.4259 | 0.2042 | 0.0506 |
| Polynomial | 0.0000<br>0.0000<br>0.0325<br>15.4486 | 0.0000<br>0.0001<br>0.1566<br>15.4486 | 0.0854 | 0.8324 |

Table 4.4: Results of example toe strategy

(From top to bottom, values in the $K_p/m$ and $K_d/m$ cells of 'Polynomial' are gains for $e^7$, $e^5$, $e^3$ and $e$ terms)

| Strategies \ Parameters | $K_p/m$ | $K_d/m$ | RMS Error | $R^2$ Metric |
|---|---|---|---|---|
| Linear (Base-line) | 12.2639 | 2.446 | 0.1048 | 0.9242 |
| Exponential | -0.0782 | 0.0113 | 0.3823 | -0.0088 |
| Polynomial | 0.0000<br>0.0044<br>0.8720<br>12.2723 | -0.0003<br>-0.0503<br>-6.2343<br>2.4830 | 0.1034 | 0.9259 |

## 4.4 Strategy Selection Statistics

The statistics results we have for all trials are shown in Figure 4.7a with data points of every strategy represented with different shapes. To compare with Stephens' work[45], we have the same stable area as in Figure 4.7b, which is the area between the two lines $v = -3p + 0.3$ and $v = -3p - 0.3$. In these equations, $v$ and $p$ represent the velocity and position of CoM. When comparing these two diagrams, we have ankle strategies and a large portion of toe strategies lying within the stable area derived by the PD control law applied to a LIPM model. However, according to the human data, this stable area seems to be larger. The distributions of this scatter plot reinforce the idea that when more intense pushes are applied and more extreme CoM position is given, it leads to the selection of more effective push recovery strategy (stepping strategies). Moreover, there exists a band where selection of different strategies is more blurred which is clearly shown by the boundary of $v = -3p + 0.3$.

Because the scatter plot containing all kinds of strategies can make it difficult to observe, we have another set of scatter plots containing each kind of strategy, shown in Figure 4.8. Based on the five separate scatter plots we can see all CoM properties of ankle strategy trials at the start of a trial are within the stable area mentioned above, and those of all stepping strategies, including toe-to-step strategy, are in the unstable region except several outliers. For toe strategy, more than half of the CoM starting points lie within the stable region while all other points are in a band area of the unstable region except minor outliers, suggesting that when using toe strategy, the system does not entirely follow the theoretical prediction because it seems to have a blurry band of extra 'stable' region added to the calculated region. The pattern seems to be contiguous to that of toe-to-step strategy while we cannot draw strong conclusion since we have not gathered sufficient data points for this strategy.

To be more explicit about the detailed numbers, we have the expectations and the absolute deviations shown in Table 4.5. When building the table, we decide to use median values as our expectations because according to Figure 4.8, a few outliers exist in each strategy type which can have a significant impact on mean values, while using median values helps deal with this issue. Therefore, the deviation is also calculated with respect to medians. To be noted that we have another well-inform and random start trials separated to form another two sub-tables for comparison ease. Moreover, scatter plots of all well-informed trails and all random start trials can be found in Figure C.5 to see the strategy selection distribution differences under the two situations.

According to the table, ankle and toe strategies are usually chosen when CoM velocity is relatively lower and CoM position close to 0 (steady-state point). When recover strategy becomes more active, the mean values of these two properties grow considerably with velocity MADs getting larger as well. Expectation values of the two direct stepping strategies are more significant in random start trials compared with well-informed trials. On the other hand, MADs of random start trials except for the ones of direct stepping trials are also larger. One potential reason is that when being well-informed before the start, participants seem to have preparations whereas random start trials force them to decide based on the push force impact.

Table 4.5: Strategy selection statistics, presenting medians and median absolute deviations of the CoM properties

| Parameters (*m, m/s*) \ Strategy Type | Ankle | Toe | Toe-to-step | One-step | Two-step |
|---|---|---|---|---|---|
| **All Trials** | | | | | |
| Median | -0.0281, 0.1222 | 0.0272, 0.1686 | 0.0887, 0.2971 | 0.0515, 0.4334 | 0.1006, 0.4068 |
| $MAD_{Median}$ | 0.0197, 0.0477 | 0.0234, 0.0420 | 0.0130, 0.0661 | 0.0501, 0.0837 | 0.0326, 0.0930 |
| **Well-informed Trials** | | | | | |
| Median | -0.0180, 0.1404 | 0.0239, 0.1712 | 0.0810, 0.2971 | 0.0887, 0.3672 | 0.0951, 0.3764 |
| $MAD_{Median}$ | 0.0219, 0.0321 | 0.0224, 0.0348 | 0.0189, 0.0233 | 0.0518, 0.0987 | 0.0287, 0.0748 |
| **Random Start Trials** | | | | | |
| Median | -0.0327, 0.1049 | 0.0314, 0.1632 | 0.0915, 0.2922 | 0.0367, 0.4734 | 0.1150, 0.4211 |
| $MAD_{Median}$ | 0.0161, 0.0518 | 0.0249, 0.0450 | 0.0126, 0.0705 | 0.0568, 0.0697 | 0.0300, 0.0676 |

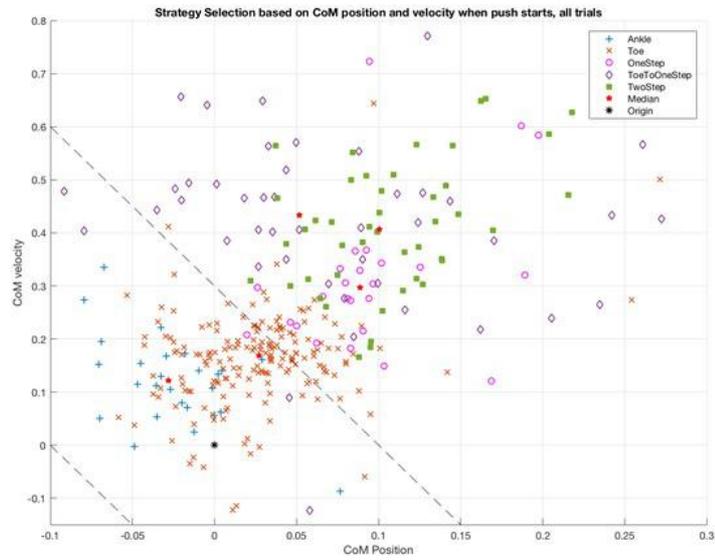

(a) Scatter plot of starting point presenting the CoM's position and velocity of all trials

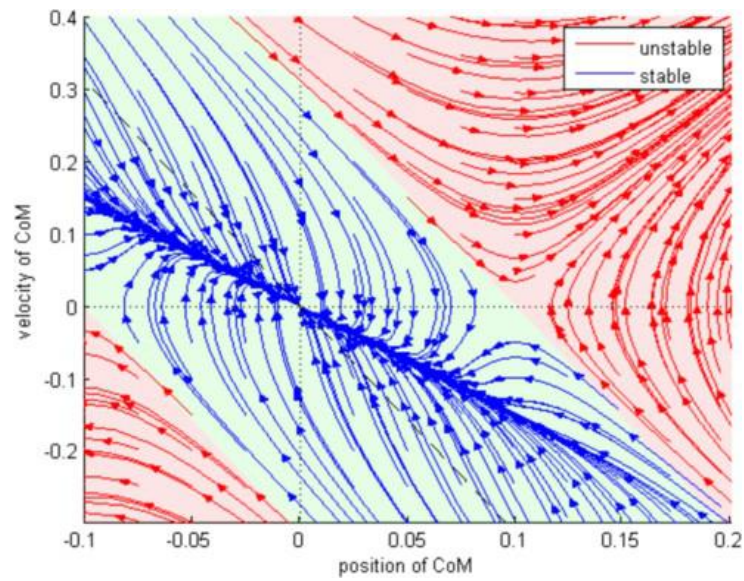

(b) CoM trajectories of feedback control created by a PD controller (with torque limits) on the ankle joint of a single inverted pendulum. *Source*: B. Stephens, 2007 [45]

Figure 4.7: Strategy selection statistics, CoM position & velocity scatter plot and trajectory plot of previous related work on humanoid push recovery
The area between the two dash lines in (a) are the same as the stable area in (b), although the axis scales are different

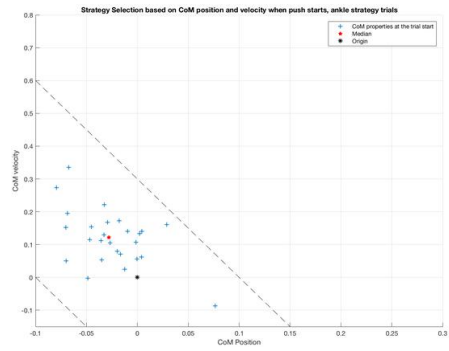
(a) Ankle strategy

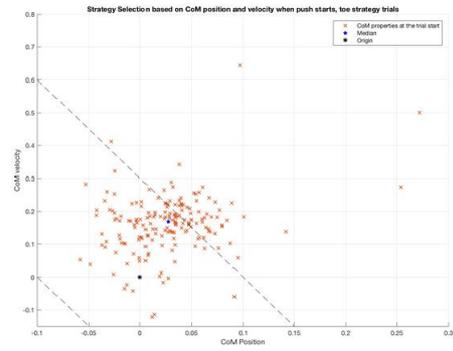
(b) Toe strategy

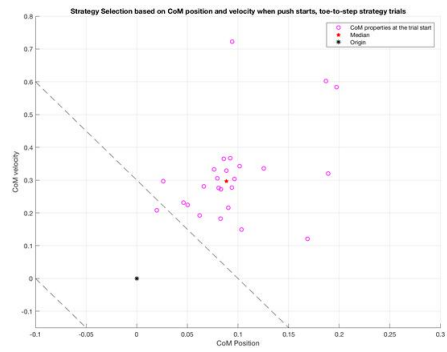
(c) Toe-to-step strategy

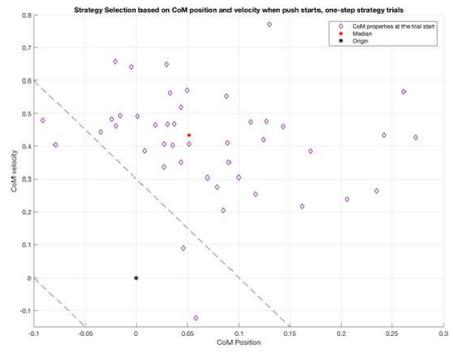
(d) One-step strategy

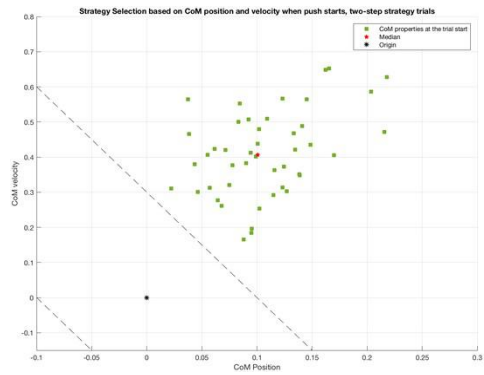
(e) Two-step strategy

Figure 4.8: Strategy selection statistics, CoM position & velocity scatter plot for all trials of every strategy

# Chapter 5

# Conclusion and Further Work

## 5.1 Conclusion

In this project, we intend to find the general control law that human beings use for push recovery. We have designed and implemented a full set of experiment to capture human data and export CoM related information to form a point mass model of the CoM. After data collection and data processing, we gain 323 trials from 7 participants in total to do strategy selection statistics and 273 trials to fit control laws for the 1-D point mass model of the CoM. We use the PID control law and its subsets to do model fitting with regularise linear regression. Moreover, our hypothesis is that PD control can describe the control relationship using the 1-D point mass model of the CoM. According to all the results and discussion outcomes presented in chapter 4, we can now conclude our project.

Within the scope of our study, control laws in all five strategies of human push recovery can be best-described by PID control when considering a set of constant gains for the whole trial. However, according to models fitted with PID control law, the integral gain is usually much smaller than the other two gains, and by comparing P and PI, as well as PD and PID control laws we strengthen our standpoint that adding the integral term only makes the model more complicated thus it becomes capable of taking more data points into accounts, including noises, to get a better fit, which is similar to over-fitting. Therefore, we reason that PD control law can be a sufficient and more reasonable control law to describe the push recovery process and the model's overall performance has an average RMS error below 0.13 and an average $R^2$ metric above 83%. According to the results from the full-trial model fitting of stepping strategies and toe-to-step strategies, it is clear that having a constant set of gain for the whole trial cannot make the model capture all trending details, which leads to our further analysis.

By segmenting a trial based on hierarchical motion primitives, we fit models on these segmentations and find the outcome fittings becomes better by having large $R^2$ metrics and smaller RMS errors on average, as well as controller domains having better descriptions of surfaces where the segmented trajectories lie on. We also find different phases of a trial having significant differences in proportional gains, which supports our hypothesis that people may set different gains for different phases of a push recovery trial, for every push recovery strategy. When fitting models on segmentations, the PD control law also fits well, and we have the same conclusion on the integral term as above when using PID control for full-trial fitting. PD control gets better performances that average RMS error is smaller than 0.1 and average $R^2$ metric hitting over 90% when fitting on segmentations. However, for toe strategy, because gravity is included in movement of dropping to stand on full-sole from standing on the tip-toe. The contaminated data in 1-D cannot help distinguishing how much of the performance is by human control and how much is by the gravity, while our model assumes full control of human beings. The same goes for the 'making a step' movement of the stepping strategies.

We also explore the error metric used by the control law. In this project, we tried both odd order polynomial error metric, and changing the linear error to exponential scale. According to the results we see that higher-order polynomials cannot improve the overall performance because higher order terms have negligible gains. Besides, changing to the exponential scale makes the performance worse, proving that sign is crucial in control models.

Finally, we have statistics of CoM's position and velocity, as well as the corresponding strategy selected when the push starts. Based on the scatter plot we find that it corresponds to the work of B. Stephens on humanoid push recovery [45]. However, we also find that human beings have a considerable blurry band of strategy selection where different strategy selections are mixed. Currently we find that the toe strategy and toe-to-step strategy can cross the calculated stable boundary. The main pattern we find is that the further the push start point is from the origin (0 *m*, 0 *m/s*), the strategy for more intensive push compensation will be chosen.

## 5.2   Critical Analysis and Further Work Suggestions

After finishing the project, we find some aspects that may need improvement. A significant part of these aspects is for data collection experiment. In the data collection phase, we use a six-camera VICION motion capture system and a MOTEK treadmill. Because of the equipment placement and our proposed method which can be checked in Figure 3.3, we have to make participants face backwards since the treadmill's belts can only move in one direction. This makes the overall motion capture unstable, and it costs us a tremendous amount of time to do the gap filling centred post-processing of raw motion data. Therefore, having more camera or more reasonable camera placement can really help in efficiency.

On the other hand, because our treadmill's total length is 1.8 *m*, we need to stop the treadmill somewhere around half of its length considering safety issues. Thus, participants have to recover from the motor's jerk and return to a normal pose within 0.9 *m* which is sometimes difficult to accomplish since usually they only have about one second to do so, and one reason for our leaving data out in further stages lies here. Plus, since we have to stop the treadmill somewhere around the middle of the treadmill, we suspect our participants get adaptive to the pattern and try to have predictive countermeasures which make our experiment data more or less biased even though we have made corresponding preparation in our design. Therefore, we suggest having a longer treadmill such that participants can have more time to recover and prepare for the treadmill's stopping, while researchers can have a broader range to stop the treadmill to have the push generation more random to the participants.

As for the data analysis stage, we use an unconstrained regularised least squares linear regression method, which gives some results having unwanted negative parameters (control gains). Although this suggests the model applied does not meet the requirement when reasoning, having a proper set of parameters may give stronger evidence. Therefore, we turn to work on the constrained version of the same regularised linear regression to have all fitted parameters to be positive. We have turned to the 'lsqlin' MATLAB function [37], but because of the time remaining, we have not completed this work. On the other hand, we have only worked on the 1-D point mass model, one of the most fundamental yet simplest models, and the project ends with the model fitting & evaluation phase. Therefore, generalising the dynamics into the 3-D case and find how the controller's behaviour varies can be worthwhile, so as using more complicated dynamic models. On the other hand, it is also worthy to extend our work to simulations with different engineering models such as the LIPM family, which is the work undertaken by C. Mcgreavy[1].

Although we have found a reasonable linear control law to describe the underlying control relationships in human push recovery with the 1-D point mass dynamics model, the real control law does not necessarily need to be linear. Therefore, some expansion of model fitting can be done with non-linear control laws such as the Sliding Mode Control [15]. Other control laws can also be applied to model fitting to get a clearer page of human push recovery exploration. Finally, our preliminary statistics of push recovery strategy selection also leaves potential human decision-making related studies in the push recovery area.

---

[1] https://www.edinburgh-robotics.org/students/christopher-mcgreavy

# Appendix A

# Experiment Set-up and Software List

## A.1 Additional Information for Experiment Set-up

To give a direct view of our experimental design, some photos showing the start and end of a trial can be found in Figure A.1. When taking the photos, we include the VICON recording interface as well for observation ease. In the figure, we can also see some camera placement in the VICON recording which may help in inferring the equipment set-up.

## A.2 Complete Program List

The following Table A.1 consists of the complete software list of this project and their main functionality we used.

Table A.1: Program List with software versions and functionality

| Program Name | Version | Functionality |
| --- | --- | --- |
| MATLAB | 2018a | Main program used for data processing and analysis. |
| VICON Nexus | 2.5 | Software of VICON motion capture system, used for recording motion data and their post-processing. |
| D-Flow | 3.16.2 | Controls the treadmill and outputs force plate data in .csv format. |
| OpenSim | 3.3 | Convert C3D marker files to generate muscle models and check these models using inverse kinematics test with corresponding trajectory files. |
| Exopt-toolbox | - | Contains OpenSim API, used for data processing and .mat dataset file generation. |
| Modified NOtoMNS Toolbox | - | Used to combine motion file and force files to generate .mot and .trc files which can be used in OpenSim. |

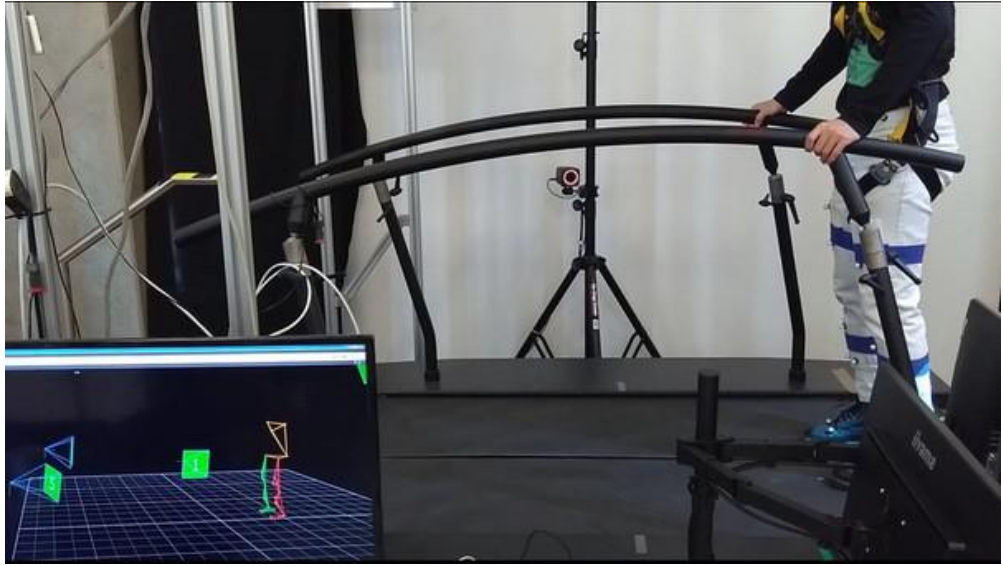

(a) The start of a trial

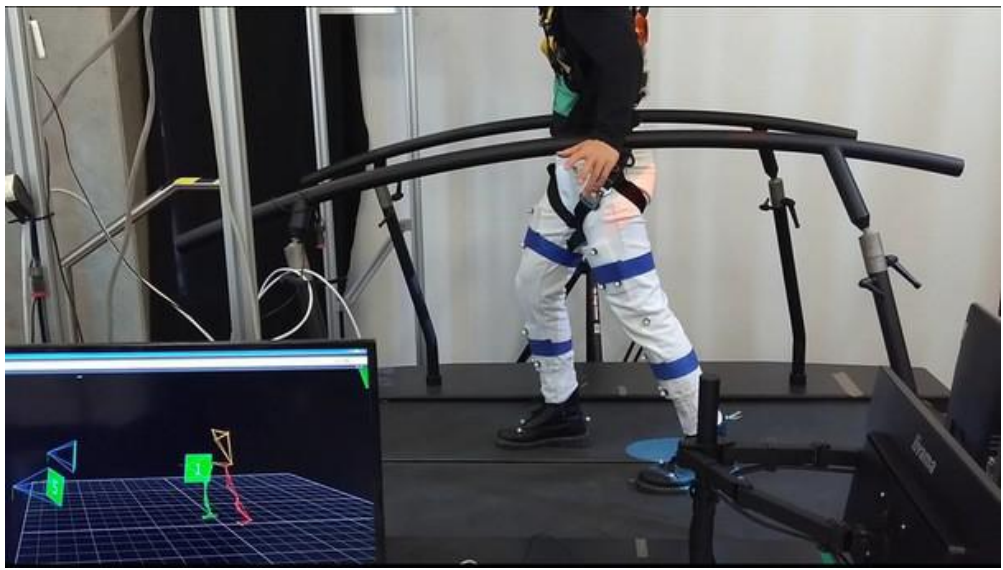

(b) The end of a trial

Figure A.1: Examples for trial start and ending

# Appendix B

# Documents for Participants

Here we attach all three documents that a participant needs to read and sign. These documents are listed in the order of experiment information sheets, clinical & experimental form and consent form. The documents begin at the next page.

Participant Information Sheet May 2018

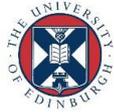

# Participant Information Sheet

**Study title: Transferring Human Push Recovery Behaviour to Robotics**

1. Aim of the study

This study aims to improve the ability of robots to recover from disturbances by external forces using data collected from humans. By applying forces to humans to see how they react, it may be possible to extract their behaviour in order to transfer onto robotic systems. The data which will be collected as part of the study will be used to develop a controller for a robot to recover from disturbances when pushed so as to not fall over.

2. Execution of the study

The data collection sessions will be conducted by researchers trained for this purpose. You will be asked to stand on a treadmill (see Figure 1a) and will be moved forwards by the treadmill, which will stop suddenly. This sudden stop will be enough to cause a force in the forward direction. You will take a small step forwards to recover your balance. To measure your joint movements accurately, you will be required to wear small reflective markers (see Figure 1b) which will be attached to the skin at pre-set points using medical tape. You will stand on the treadmill and be moved forwards at a number of different speeds. Speeds will be gradually increased to find the one which causes a small step forward but does not cause too much instability. In addition, you will be asked to fill in a consent form and a clinical and experimental data form.

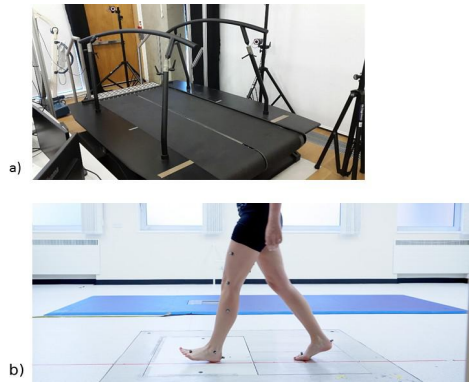

*Figure 1: (a) Treadmill, (b) Participant wearing reflective markers*

School of Informatics, 10 Crichton Street, Edinburgh, (City of) Edinburgh, EH8 9AB

3. Your participation

The data acquisition session will last for approximately 60 minutes and will take place in room G.03 of the Informatics Forum, University of Edinburgh. Participation in this study is entirely voluntary. You can refuse to take part or withdraw from the study at any time without having to give a reason. Such a decision has no adverse implications for you.

4. Risk assessment

Your participation to this study involves risk that are as low as reasonably possible. You will be wearing a harness at all times whilst walking on the treadmill, this will ensure that you cannot fall.

5. Privacy

All data acquired will be treated confidentially. The data might be disclosed anonymously to third parties for the purposes of the study. Your personal information will be stored separately to ensure data protection.

6. Contact

If you have any questions or require further information, please do not hesitate to contact Mr. Christopher McGreavy (c.mcgreavy@ed.ac.uk).

Clinical and Experimental Form                                                                         May 2018

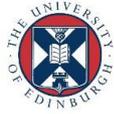

# Clinical and Experimental Form

**Study title: Transferring Human Push Recovery Behaviour to Robotics**

**To be completed by the researcher:**

| Subject Number | |
|---|---|
| DoB | |
| Mass | |
| Height | |
| Date | |
| Time | |

**To be completed by the participant:**

| Given name | |
|---|---|
| Family name | |
| Phone number | |
| E-mail address | |

School of Informatics, 10 Crichton Street, Edinburgh, (City of) Edinburgh, EH8 9AB

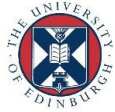

# Informed Consent Form

**Study title: Transferring Human Push Recovery Behaviour to Robotics**

1. I confirm that I have read and understood the Participant Information Sheet for the above study and there is no reason I should not take part. I have had the opportunity to consider the information and ask questions, and have had these answered satisfactorily.
2. I understand that my participation is entirely voluntary and I am free to withdraw at any time without giving a reason.
3. I certify that I have been informed that the data collected during the study will be shared with the scientific community in respect of anonymity, only researchers directly involved with the data acquisition and storage will have direct knowledge of my identity, and they will be bound by professional secrecy.
4. I understand that data collected in this study will be used for the project named above and will also form a part of longer term studies into human movement. Data will be disposed of once these studies are complete.
5. I have been made aware of the risk assessment carried out for the treadmill and the safety measure that are in place to minimize risk.
6. I agree to take part in this study.

………………………………………        ………………………….         ………………………………….
Name of participant                        Date                              Signature

………………………………………        ………………………….         ………………………………….
Name of researcher                        Date                              Signature

School of Informatics, 10 Crichton Street, Edinburgh, (City of) Edinburgh, EH8 9AB

# Appendix C

# Additional Support Figures

This appendix contains figures supporting the arguments in the main body. Since putting all these figures in the main body is trivial and unnecessary, we move them here in a separate group. They are also referred to in the main body when related arguments is being presented.

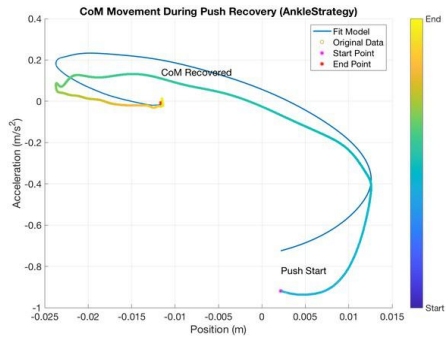
(a) Position-acceleration view, ankle strategy

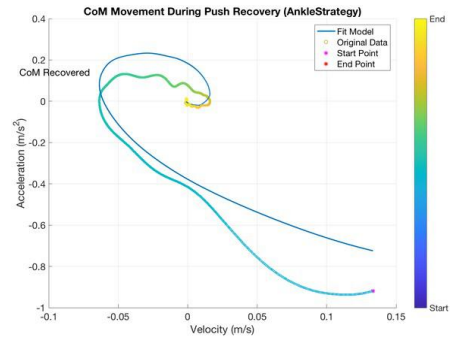
(b) Velocity-acceleration view, ankle strategy

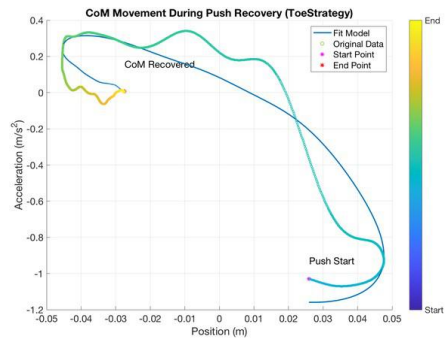
(c) Position-acceleration, toe strategy

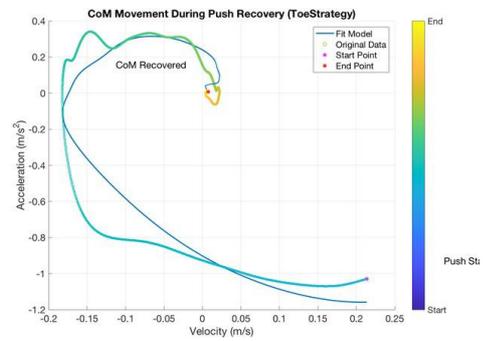
(d) Velocity-acceleration view, toe strategy

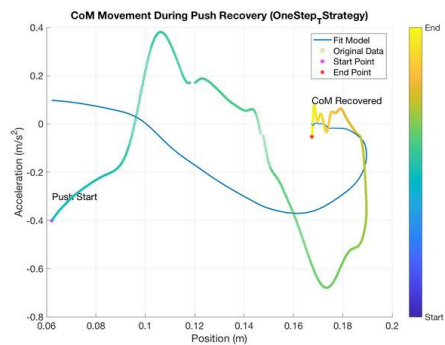
(e) Position-acceleration view, toe-to-step strategy

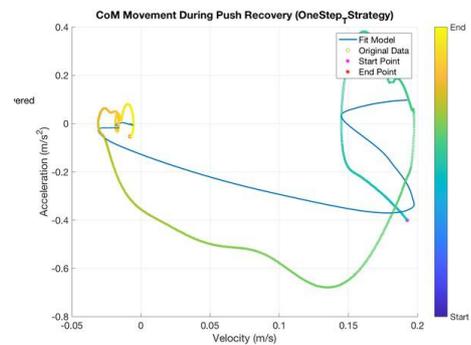
(f) Velocity-acceleration view, toe-to-step strategy

Figure C.1: Example trials of all strategies, fit with PD control law

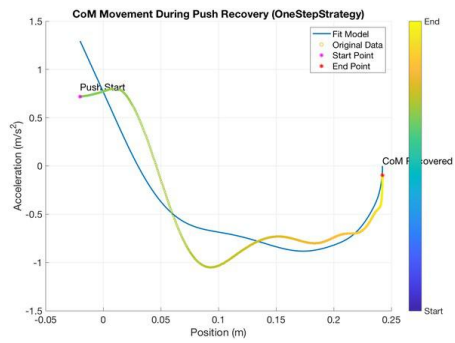

(g) Position-acceleration view, one-step strategy

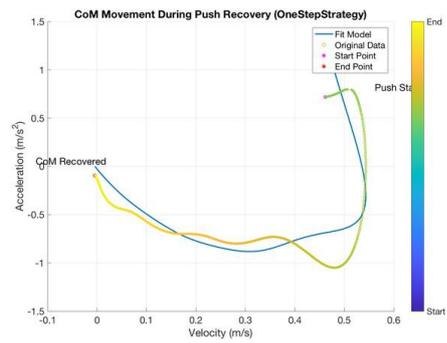

(h) Velocity-acceleration view, one-step strategy

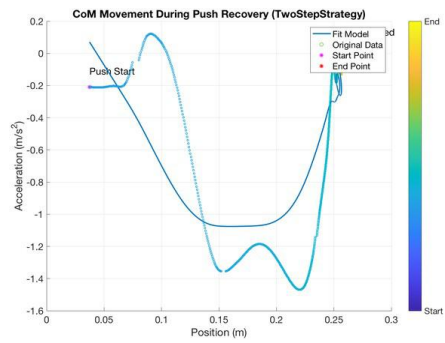

(i) Position-acceleration, two-step strategy

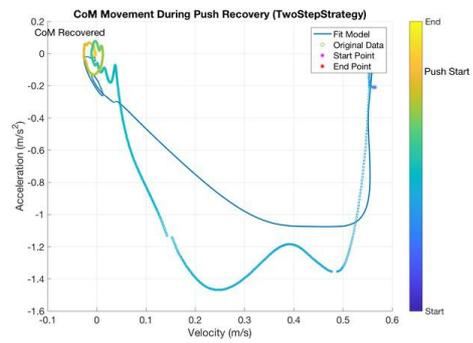

(j) Velocity-acceleration view, two-step strategy

Figure C.1: Example trials of all strategies, fit with PD control law (cont.)

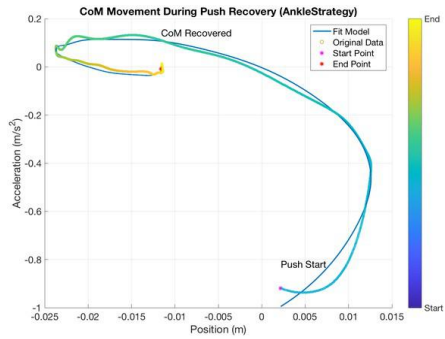
(a) Position-acceleration view, ankle strategy

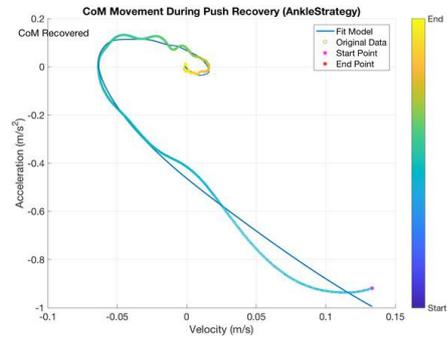
(b) Velocity-acceleration view, ankle strategy

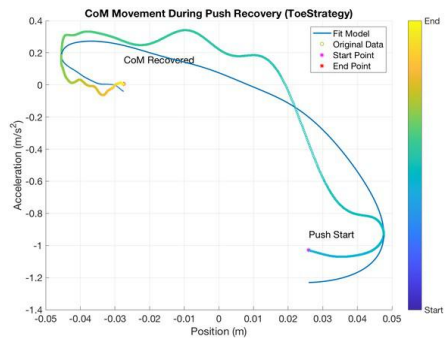
(c) Position-acceleration, toe strategy

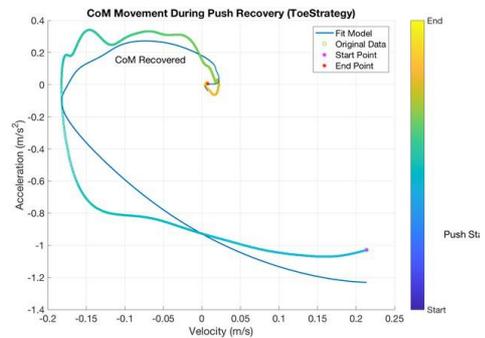
(d) Velocity-acceleration view, toe strategy

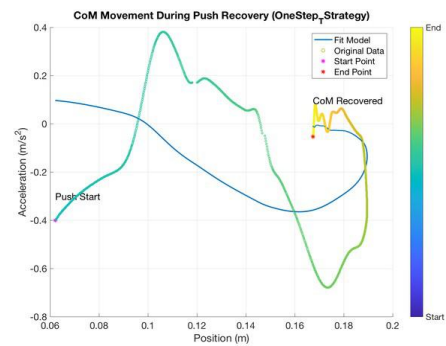
(e) Position-acceleration view, toe-to-step strategy

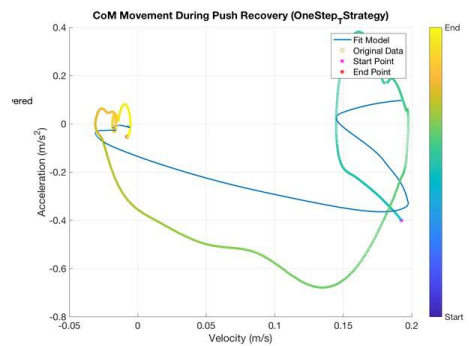
(f) Velocity-acceleration view, toe-to-step strategy

Figure C.2: Example trials of all strategies, fit with PID control law

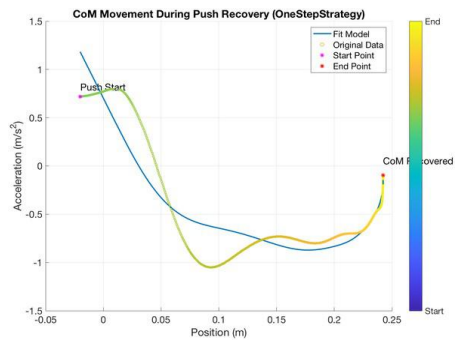
(g) Position-acceleration view, one-step strategy

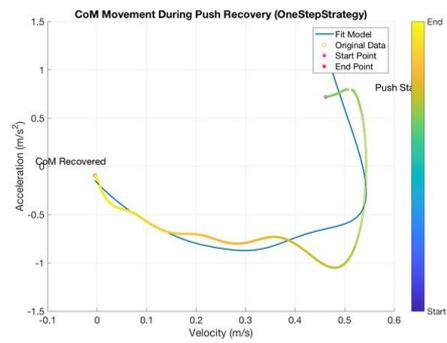
(h) Velocity-acceleration view, one-step strategy

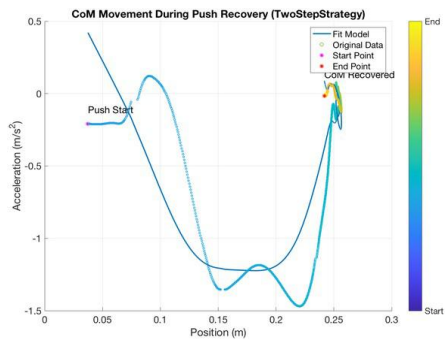
(i) Position-acceleration, two-step strategy

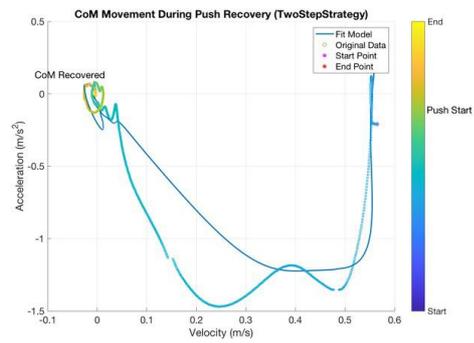
(j) Velocity-acceleration view, two-step strategy

Figure C.2: Example trials of all strategies, fit with PID control law (cont.)

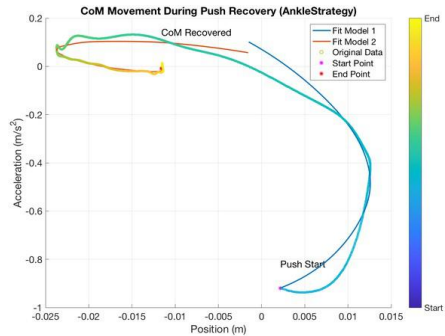
(a) Position-acceleration view, ankle strategy

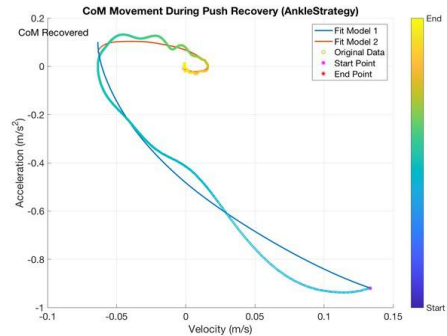
(b) Velocity-acceleration view, ankle strategy

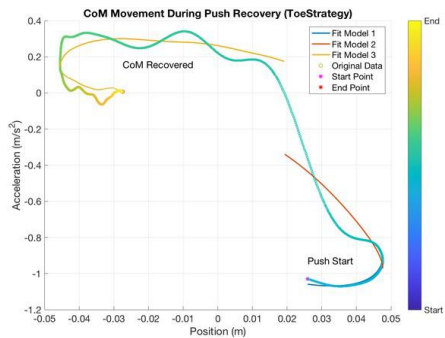
(c) Position-acceleration, toe strategy

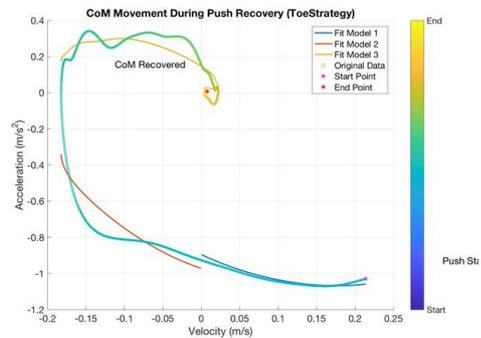
(d) Velocity-acceleration view, toe strategy

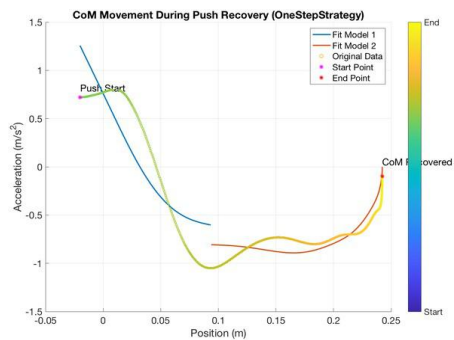
(e) Position-acceleration view, one-step strategy

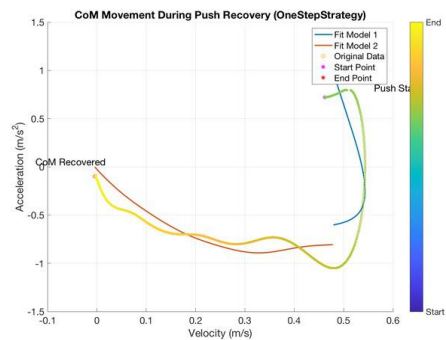
(f) Velocity-acceleration view, one-step strategy

Figure C.3: Example trials of the three selected strategies, fit with PD control law

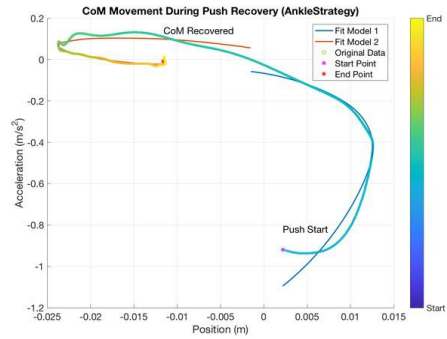
(a) Position-acceleration view, ankle strategy

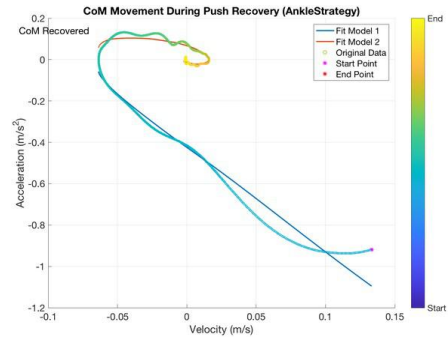
(b) Velocity-acceleration view, ankle strategy

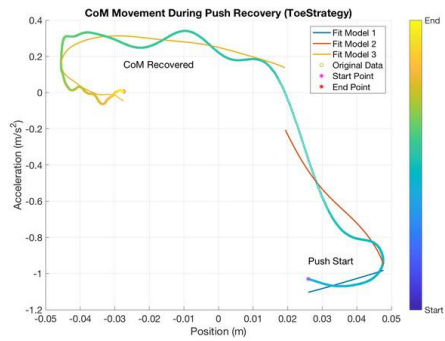
(c) Position-acceleration, toe strategy

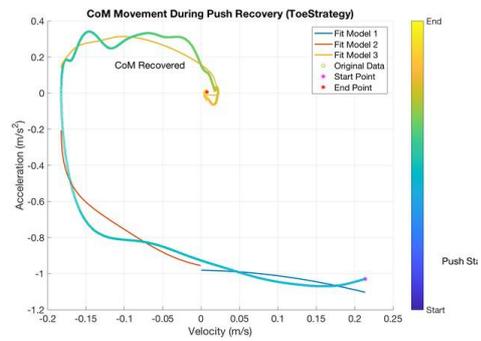
(d) Velocity-acceleration view, toe strategy

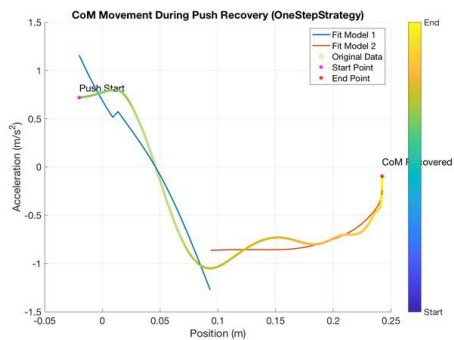
(e) Position-acceleration view, one-step strategy

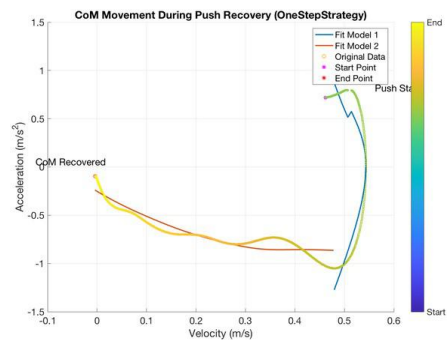
(f) Velocity-acceleration view, one-step strategy

Figure C.4: Example trials of the three selected strategies, fit with PID control law

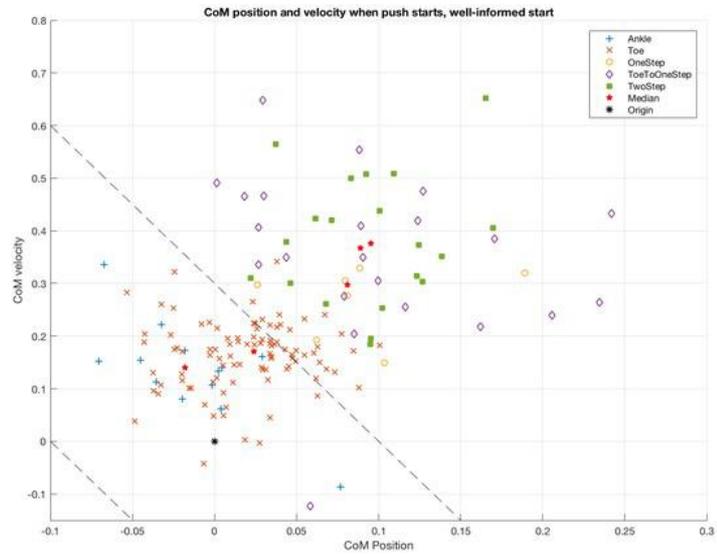

(a) Well-informed trials

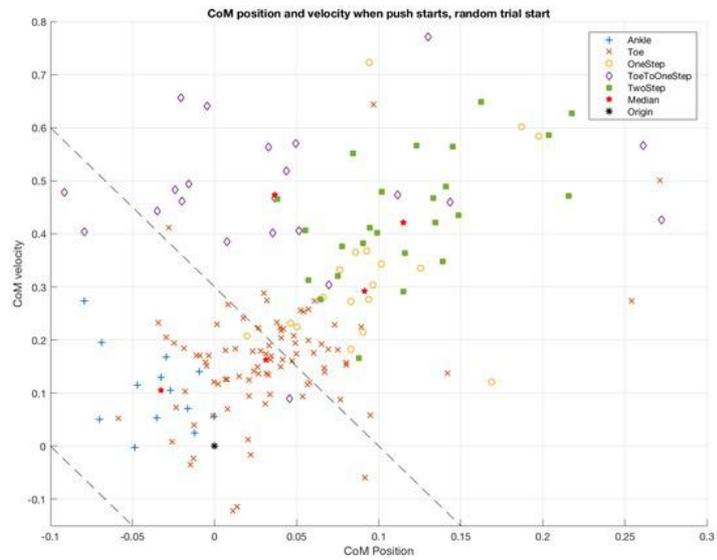

(b) Random start trials

Figure C.5: Strategy selection statistics, CoM acceleration & velocity scatter plot and trajectory plot of related previous work on humanoid push recovery